\UseRawInputEncoding

\documentclass[journal]{IEEEtran}
\ifCLASSINFOpdf
\else
\fi

\usepackage{times}
\usepackage{epsfig}
\usepackage{graphicx}
\usepackage{amsmath}
\usepackage{amssymb}
\usepackage{subfiles}
\usepackage{multirow}
\usepackage{subcaption}
\usepackage{tablefootnote}

\usepackage{balance}
\usepackage{cite}

\usepackage[colorlinks,linkcolor=red]{hyperref}
\usepackage{makecell}
\usepackage[table ]{ xcolor}
\definecolor{mygray}{gray}{.9}
\definecolor{white}{gray}{1}
\usepackage{color}
\usepackage{enumitem}
\usepackage{url}
\usepackage[square,sort&compress,numbers]{natbib}
\begin{document}
%
\title{Anti-UAV: A Large Multi-Modal Benchmark for UAV Tracking}
%
%
%

\author{Nan Jiang$^{\dagger}$, Kuiran Wang$^{\dagger}$, Xiaoke Peng$^{\dagger}$, Xuehui Yu, Qiang Wang, Junliang Xing, Guorong Li, \\ Jian Zhao$^{\ddagger}$, Guodong Guo, \textit{Senior Member, IEEE}, and Zhenjun Han$^{\ddagger}$ 
\thanks{$^{\dagger}$\ Equal contribution.}
\thanks{$^{\ddagger}$\ These authors jointly supervised this work: Zhenjun Han and Jian Zhao.} 
\thanks{N. Jiang, K. Wang, X. Peng, X. Yu, G. Li and Z. Han are with University of Chinese Academic of Sciences (UCAS), Beijing, 101408 China. E-mail: \{jiangnan18, wangkui\-ran19, pengxiaoke19, yuxuehui17\}@mails.ucas.ac.cn, \{liguorong, hanzhj\}@ucas.ac.cn.}
\thanks{Q. Wang and J. Xing is with Institute of Automation, Chinese Academy of Sciences (CASIA), and also with  University of Chinese Academic of Sciences (UCAS), Beijing, China. E-mail: \{qiang.wang, jlxing\}@nlpr.ia.ac.cn.}
\thanks{J. Zhao is with Institute of North Electronic Equipment, Beijing, China. Homepage: https://zhaoj9014.github.io/. E-mail: zhaojian90@u.nus.edu.}
\thanks{G. Guo is with Institute of Deep Learning, Baidu Research and National
Engineering Laboratory for Deep Learning Technology and Application. Email: guoguodong01@baidu.com.}}

%
%

\maketitle
\begin{abstract}
    Unmanned Aerial Vehicle (UAV) offers lots of applications in both commerce and recreation. Therefore, perception of the status of UAVs is crucially important. In this paper, we consider the task of tracking UAVs, providing rich information such as location and trajectory.
    To facilitate research on this topic, we introduce a new benchmark, referred to as Anti-UAV, opening up a promising direction for UAV tracking in a long distance with more than 300 video pairs containing over 580k manually annotated bounding boxes. Furthermore, the advancement of addressing research challenges in Anti-UAV can help the design of anti-UAV systems, leading to better surveillance of UAVs.
    Accordingly, a simple yet effective approach named dual-flow semantic consistency (DFSC) is proposed for UAV tracking. Modulated by the semantic flow across video sequences, the tracker learns more robust class-level semantic information and obtains more discriminative instance-level features. Experiments show the significant performance gain of our proposed approach over state-of-the-art trackers, and the challenging aspects of Anti-UAV. The Anti-UAV benchmark and the code of the proposed approach will be publicly available at \url{https://github.com/ucas-vg/Anti-UAV}.

\end{abstract}


%
\IEEEpeerreviewmaketitle

\section{Introduction}
\IEEEPARstart{O}bject tracking is to locate an object across a series of video frames \cite{DBLP:journals/tist/LiHSZDH13}\cite{ DBLP:journals/pami/SmeuldersCCCDS14}.
It is widely used in video surveillance \cite{DBLP:journals/pami/HaritaogluHD00}, maritime rescue \cite{DBLP:journals/ijon/BrunettiBTB18}\cite{DBLP:conf/wacv/YuGJYH20}
and self-driving cars \cite{DBLP:conf/cvpr/ChangLSSBHW0LRH19}.
Recently, the accessibility and popularity of Unmanned Aerial Vehicle (UAV) for commercial and recreational use have significantly surged.
It has a wide range of applications, such as autonomously landing \cite{DBLP:journals/arobots/LinGL17}, target tracking and following \cite{DBLP:conf/iros/ChengLZGL17}.Behind these practical applications, it is crucial to monitor the operation status of UAV, including locations and trajectory. To this end, this article focuses on tracking UAV.

In literature, most of object trackers are based on RGB information \cite{DBLP:conf/eccv/BertinettoVHVT16,DBLP:conf/cvpr/LiYWZH18,DBLP:conf/cvpr/LiWWZXY19,DBLP:conf/cvpr/Wang0BHT19, DBLP:conf/cvpr/DanelljanBKF19}. 
However, when in the low light conditions, these trackers might not be able to find useful cues, leading to unreliable results.
To alleviate this, some works consider using infrared (IR) images for object tracking \cite{DBLP:journals/tip/VenkataramanFHFZY12}\cite{DBLP:conf/cvpr/BergAF16}.
The crucial drawback of infrared images is that it usually has a low resolution, providing inadequate information for trackers.
Based on the above analysis, we think it is important to fuse information from visible RGB and IR images for tracking UAVs. Specifically, leveraging multi-modal information is beneficial for learning accurate and robust UAV trackers \cite{DBLP:conf/cvpr/WangZ0H0L18}\cite{DBLP:journals/chinaf/LiuS12}. For this purpose, a multi-modal dataset is required to study the task of tracking UAV.

To the best knowledge, there is no multi-modal UAV tracking benchmark for UAV. In the community, the tracking datasets mainly contain general objects (\ e.g., car and person). To mitigate this, we introduce a new dataset, Anti-UAV, to facilitate the research on UAV tracking. 
Anti-UAV contains high-quality and high-definition video sequences of both RGB and IR. Each sequence is annotated with bounding boxes, attributes, and flags indicating whether the target object exists.
It is worth noting that RGB and IR video sequences are \emph{paired} in Anti-UAV. Thus, the proposed dataset supports both single-modal and multi-modal UAV tracking. 

Furthermore, based on the fact that Anti-UAV has only one category, this work proposes a novel training strategy named dual-flow semantic consistency (DFSC), containing class-level semantic modulation (CSM) stage and instance-level semantic modulation (ISM) stage for UAV tracking. On the one hand, all different sequences share one trait: the labeled objects in them are only UAVs, which means the network can utilize the cross-sequence features. The tracker in the CSM stage adopts class-level semantic modulation to retrieve anchors that may contain UAV as far as possible to reduce the intraclass differences. On the other hand, the tracker pays more attention to the fine-grained instance-level features to distinguish between the real UAV instance of the current tracking sequence and the similar distractors in the ISM stage. Since the DFSC method only works in training, it will not affect the calculation and time consumption in the inference time.

The main contributions of this work can be summarized as:

\begin{itemize} 
\item A multi-modal dataset, named Anti-UAV, is constructed. The Anti-UAV dataset consists of 318 RGB-T video pairs with ground-truth in different scenes, where ``RGB'' and ``T" represent visible light and thermal infrared, respectively. It has been released publicly for academic research.

\item To better evaluate the tracker's performance, an evaluation metric that focuses on judging the state of UAV is given. Experimental evaluation and analysis of more than 40 trackers on Anti-UAV are provided. 

\item A novel training strategy, named dual-flow semantic consistency (DFSC), is proposed specifically for UAV tracking. DFSC guarantees the tracker obtain robust and representative feature after CSM and ISM, finally significantly outperforms the baselines.

\end{itemize}

Served as a large-scale multi-modal benchmark, Anti-UAV drives the future research on the frontiers of tracking UAVs in the wild. With the above innovations and contributions, we have organized the CVPR 2020 Workshop on the $1^{st}$ Anti-UAV Challenge. These contributions together significantly benefit the community. 

\section{Related Work}

\begin{figure}[t]
\begin{center}

 \includegraphics[width=1\linewidth]{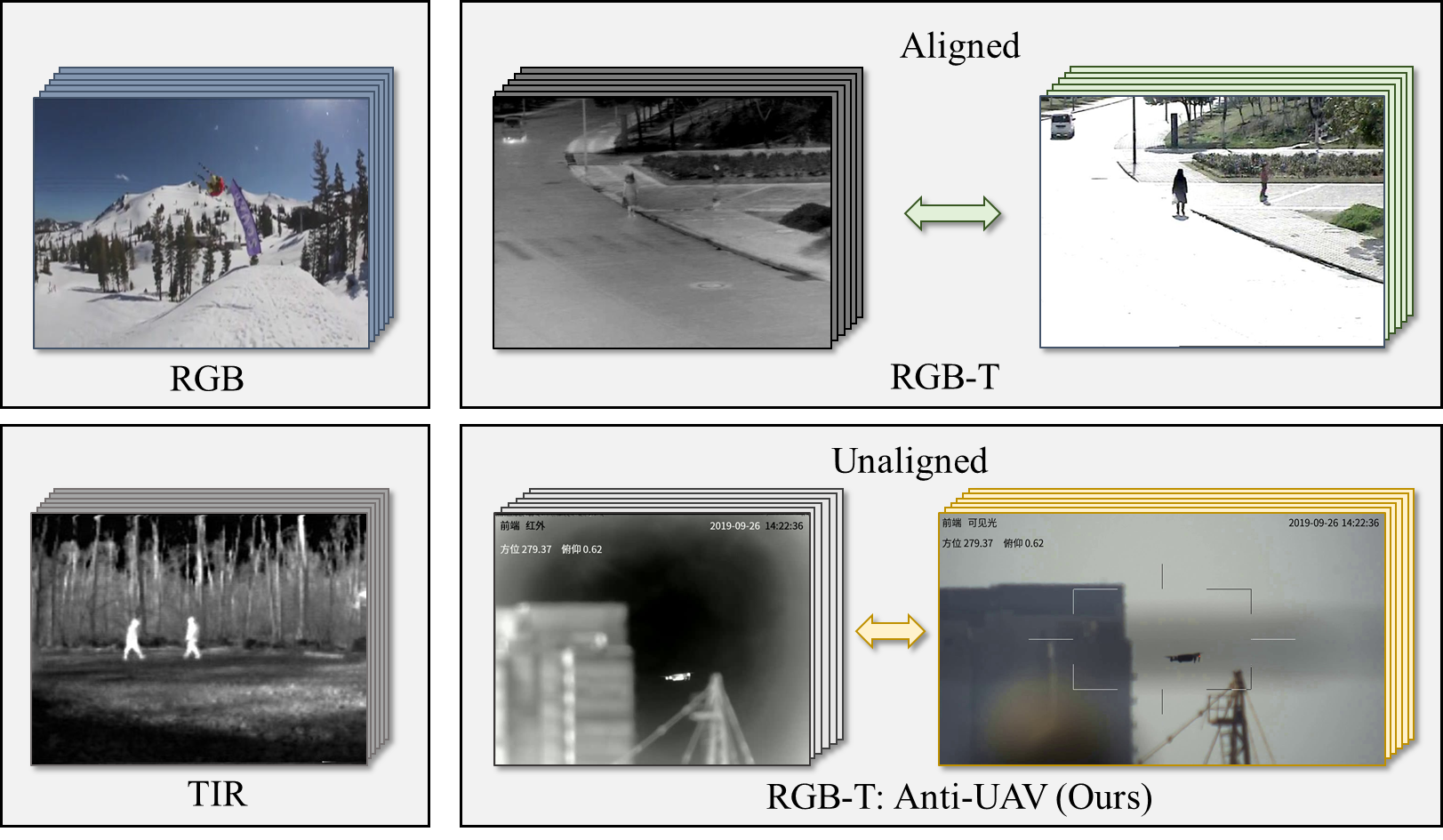}
\end{center}
 \vspace{-3mm}
 \caption{Overview of tracking datasets. Unlike other RGB-T datasets, Anti-UAV is composed of unaligned video pairs.}
 \vspace{-3mm}
\label{fig:summary}
\label{fig:dataset}
\end{figure}

In the following, an overview of the various researches on object tracking is given in Fig. \ref{fig:dataset}. The related tasks are grouped into RGB, TIR, and RGB-T based datasets as shown in Tab. \ref{tab:pro_pro}. Then we analyze the similarities and differences between some SOTA trackers and the tracker proposed in this paper.

\vspace{-3mm}
\subsection{Tracking Dataset}\label{track}

\noindent\textbf{RGB Tracking Dataset.}
There are several datasets for RGB object tracking in Tab. \ref{tab:pro_pro}. 
In OTB \cite{DBLP:conf/cvpr/WuLY13}\cite{DBLP:journals/pami/WuLY15} and TC128 \cite{DBLP:journals/tip/LiangBL15}, each frame of a video sequence is annotated with 11 different properties and vertical borders while 314 video sequences with 14 attributes make up ALOV++ \cite{DBLP:journals/pami/SmeuldersCCCDS14}. VOT \cite{DBLP:conf/iccvw/KristanLMFPZVHL17}\cite{ DBLP:conf/iccvw/KristanBZRGBDDN19} with 60 video sequences 
, introduces the rotary bounding box, and studies the object tracking annotation extensively.

\setlength{\tabcolsep}{12.3pt}
\begin{table*}
 \centering
  \caption{A comparison of Anti-UAV with other single object tracking (SOT) datasets in terms of the number of video sequences, bounding boxes, and attributes. Anti-UAV is much larger than most RGB-T tracking datasets. In the meantime, Anti-UAV is specifically for UAV tracking and provides corresponding training set.} 
 \begin{tabular}{c|c||c|c|c|c|c|c|c} 
 \hline\Xhline{1.2pt}
 \multicolumn{2}{c||}{\multirow{2}{*}{\textbf{Dataset}}} &
 \multicolumn{2}{c|}{\textbf{Total}} & \multicolumn{2}{c|}{\textbf{Train}} &
 \multicolumn{2}{c|}{\textbf{Test}} & \multirow{2}{*}{\textbf{Attribute}} \\
 \cline{3-8}
 \multicolumn{2}{c||}{}&Sequences & Bboxes & Sequences & Bboxes & Sequences & Bboxes & \\
 \hline\Xhline{1.2pt}
 \rowcolor{mygray}\cellcolor{white} & OTB2013 \cite{DBLP:conf/cvpr/WuLY13} & 50& 29.4k& -& - & 50& 29.4k &11 \\
 & OTB2015 \cite{DBLP:journals/pami/WuLY15}& 100& 59k& -& - & 100& 59k &11 \\
 \rowcolor{mygray}\cellcolor{white}&VOT2014 \cite{DBLP:conf/eccv/KristanPLMCNVFL14}&25&10k&-&-&25&10k&5\\
 &VOT2017 \cite{DBLP:conf/iccvw/KristanLMFPZVHL17}&60&21k&-&-&60&21k&5\\
 \rowcolor{mygray}\cellcolor{white}&VOT2019 \cite{DBLP:conf/iccvw/KristanBZRGBDDN19}&60&19.9k&-&-&60&19.9k&5\\
 &ALOV++ \cite{DBLP:journals/pami/SmeuldersCCCDS14}& 314&16k&-&-&314&16k&14\\
 \rowcolor{mygray}\cellcolor{white}&TC128 \cite{DBLP:journals/tip/LiangBL15}&128&55k&-&-&128&55k&11\\
 &NUS\_PRO\cite{DBLP:journals/pami/LiL0YY16}&365&135k&-&-&365&135k&12\\
 \rowcolor{mygray}\cellcolor{white}&OxUxA \cite{DBLP:conf/eccv/ValmadreBHTVSTG18}&366&155k&-&-&366&155k&6\\
 &UAV123\cite{DBLP:conf/eccv/MuellerSG16}&123&113k&-&-&123&113k&12\\
 \rowcolor{mygray}\cellcolor{white}&UAV20L\cite{DBLP:conf/eccv/MuellerSG16}&20&59k&-&-&20&59k&12\\
 &Nfs \cite{DBLP:conf/iccv/GaloogahiFHRL17}&100&38k&-&-&100&38k&9\\
 \rowcolor{mygray}\cellcolor{white}&LaSOT \cite{DBLP:conf/cvpr/FanLYCDYBXLL19}&1.4k&3.3M&1.1k&2.8M&280&685k&14\\
 &TrackingNet \cite{DBLP:conf/eccv/MullerBGAG18}&31k&14M&30k&14M&511&226k&15\\
 \rowcolor{mygray}\cellcolor{white}\multirow{-16}{*}{\textbf{RGB}} &GOT-10k \cite{DBLP:journals/corr/abs-1810-11981}&10k&1.5M&9.3k&1.4M&420&56k&6\\\Xhline{1.2pt}
 \hline\hline\Xhline{1.2pt}
  &OSU-T \cite{DBLP:conf/wacv/DavisK05}&10&0.2k&-&-&10&0.2k&-\\
 \rowcolor{mygray}\cellcolor{white}&PDT-ATV \cite{DBLP:conf/icra/PortmannLCS14}&8&4k&-&-&8&4k&-\\
 &BU-TIV \cite{DBLP:conf/cvpr/WuFTB14}& 16&60k&-&-&16&60k&-\\
 \rowcolor{mygray}\cellcolor{white}&ASL-TID \cite{DBLP:conf/icra/PortmannLCS14}&9&4.3k&-&-&9&4.3k&-\\
 &TIV \cite{DBLP:conf/cvpr/WuFTB14}& 16& 63k& -& - & 16& 63k&- \\
 \rowcolor{mygray}\cellcolor{white}&LTIR \cite{DBLP:conf/iccvw/FelsbergBHAKMLC15}&20&11.2k&-&-&20&11.2k&5\\
 &VOT-TIR16 \cite{DBLP:conf/eccv/FelsbergKMLPHBE16}&25&14k&-&-&25&14k&10\\
 \rowcolor{mygray}\cellcolor{white}&PTB-TIR \cite{DBLP:journals/tmm/LiuHLZ20}&60&30k&-&-&60&30k&9\\
 \multirow{-10}{*}{\textbf{TIR}}&LSOTB-TIR \cite{DBLP:conf/mm/0001L0LLZYLYFZ20} & 1400&606k&1280&524k&120&82k&12\\\Xhline{1.2pt}
 \hline\hline \Xhline{1.2pt}
 \rowcolor{mygray}\cellcolor{white} & OSU-CT \cite{DBLP:journals/cviu/DavisS07} & 6& 17k& -& - & 6& 17k&-\\
 &LITIV \cite{DBLP:journals/cviu/TorabiMB12}&9&6.3k&-&-&9&6.3k&-\\
 \rowcolor{mygray}\cellcolor{white}&GTOT \cite{DBLP:journals/tip/LiCHLTL16}&50&15.8k&-&-&50&15.8k&7\\
 &RGBT210 \cite{DBLP:conf/mm/LiZLZT17}&210&210k&-&-&210&210k&12\\
 \rowcolor{mygray}\cellcolor{white}&RGBT234 \cite{DBLP:journals/pr/LiLLZT19}&234&233.8k&-&-&234&233.8k&12\\
 \multirow{-6}{*}{\textbf{RGB-T}}&\textbf{Anti-UAV (Ours)}&318&585.9k&160&294.4k&91&168.4k&7\\

 \Xhline{1.2pt}
 \end{tabular}
 \label{tab:pro_pro}
 \vspace{-3mm}
\end{table*}

More recently, TrackingNet \cite{DBLP:conf/eccv/MullerBGAG18} and LaSOT \cite{DBLP:conf/cvpr/FanLYCDYBXLL19} offer large-scale tracking datasets. TrackingNet selects about 30,000 videos to form the training subset from YouTube-BB \cite{DBLP:conf/cvpr/RealSMPV17}. As for the evaluation subset, it collects 511 videos whose category distribution is similar to the training subset. LaSOT collects and annotates 14,000 videos manually. GOT-10k \cite{DBLP:journals/corr/abs-1810-11981} covers a broader range of object categories with an one-shot evaluation protocol to avoid bias in evaluating the seen class.

\noindent\textbf{TIR Tracking Dataset.}
Composed of 16 sequences of over 60K frames with high resolution, BU-TIV \cite{DBLP:conf/cvpr/WuFTB14} is intended for various TIR visual tasks including tracking, group motion estimation and counting. LTIR \cite{DBLP:conf/iccvw/FelsbergBHAKMLC15} is the first standard TIR object tracking benchmark with 20 sequences, 6 target classes, and an evaluation toolkit. VOT-TIR16 \cite{DBLP:conf/eccv/FelsbergKMLPHBE16}, an extension of VOT-TIR15 \cite{DBLP:conf/iccvw/FelsbergBHAKMLC15}, contains 25 sequences and 8 object classes, making it more challenging than VOT-TIR15. VOT-TIR16 has six challenging subsets used to evaluate specific attributes of the tracker. Lately, LSOTB-TIR \cite{DBLP:conf/mm/0001L0LLZYLYFZ20}, a large-scale and highly diverse TIR object tracking benchmark, is constructed with 1400 TIR sequences over 600k frames.

\noindent\textbf{RGB-T Tracking Dataset.}
The OSU-CT dataset \cite{DBLP:journals/cviu/DavisS07} contains 6 pairs of RGB-T video sequences with limited size and low diversity. 
RGBT210 \cite{DBLP:conf/mm/LiZLZT17}, consists of 210 RGB-T video pairs recorded on a mobile platform, which enriches the diversity of the dataset. 
Li \textit{et al.} \cite{DBLP:journals/pr/LiLLZT19} collects a total of 234 video pairs to construct a large RGB-T tracking dataset named RGBT234 including more challenging RGB-T video, baseline algorithms, attributes, and evaluation metrics. 

Compared with the above datasets, it is worth noting that the video sequence pairs in Anti-UAV are not aligned, which is more challenging than other RGB-T datasets.

\vspace{-4mm}
\subsection{Trackers}

\noindent\textbf{RGB Based Trackers.}
Most traditional trackers mainly base on filtering 
\cite{DBLP:conf/cvpr/BolmeBDL10,DBLP:conf/eccv/HenriquesCMB12,DBLP:conf/cvpr/BertinettoVGMT16,DBLP:journals/tcsv/HanWY20,DBLP:conf/cvpr/DanelljanKFW14,DBLP:journals/pr/HanJZYL11,DBLP:journals/pami/HenriquesC0B15}. In the initialization stage, a discriminative filter is trained based on the minimum mean square error of the result according to the first frame sample. Then, the filter finds the target's position through the response map.

Despite some advantages of correlation filter in precision and efficiency, there are still some drawbacks. The performance of most existing trackers relies on an appropriate search area. In addition, the limitations of negative samples will also limit the ability of the filter to distinguish the background.

Moreover, neural networks are introduced to object tracking. The first attempt to apply deep learning to tracking task is made by Wang \textit{et al.}, who also creatively proposes the deep target tracking framework \cite{DBLP:conf/nips/WangY13}.
Since then, the research on the deep tracker has been widely carried out. From the aspect of structure, these methods can be divided into the methods based on pre-training network combined with correlation filter \cite{DBLP:conf/iccvw/DanelljanHKF15,DBLP:conf/iccv/MaHYY15,DBLP:conf/eccv/DanelljanRKF16,DBLP:conf/cvpr/DanelljanBKF17,DBLP:conf/cvpr/ValmadreBHVT17, 7752914}, 
siamese-based network \cite{DBLP:conf/eccv/BertinettoVHVT16,DBLP:conf/cvpr/LiYWZH18,DBLP:conf/cvpr/LiWWZXY19,DBLP:conf/cvpr/Wang0BHT19, DBLP:journals/tmm/TianLLLY21}, convolutional neural network \cite{DBLP:conf/eccv/JungSBH18,DBLP:conf/cvpr/NamH16,DBLP:journals/tmm/ZhaKLZ20,DBLP:journals/tamd/WangZWY20,DBLP:journals/ijon/LuNMY19,DBLP:journals/tcsv/HanWY20}, and so on \cite{DBLP:journals/tmm/LiangWKWF18, DBLP:journals/tmm/WangYWZ19, 7173057}. 

\noindent\textbf{TIR Based Trackers.}
Since the deep learning method has achieved great success in visual tracking, some works begin to introduce convolutional neural network (CNN) to improve the performance of TIR tracker.

MCFT \cite{DBLP:journals/kbs/LiuLHZC17} integrates deep features of VGGNet
and correlation filter 
into a TIR tracker. LMSCO \cite{DBLP:conf/icpr/GaoMSLWX18} makes a combination of appearance features and motion features to construct a TIR target tracking. 
As a multi-level similarity model, MLSSNet \cite{9138763} is proposed for robust TIR tracking under a Siamese Network. Based on the particle filter framework for TIR target tracking, a mask sparse representation deep appearance model \cite{DBLP:journals/remotesensing/LiPCHQP19} is proposed to replace the previous random sampling for searching useful candidates.

\noindent\textbf{RGB-T Based Trackers.} 
As thermal infrared sensors are increasingly popular, the number of RGB-T tracking datasets increases. Accordingly, more and more works \cite{9340007} devote their efforts to study target tracking specifically for RGB-T datasets.

Wu \textit{et al.} \cite{DBLP:conf/fusion/WuBCBL11} use the spliced image patch of RGB and heat source to represent each sample sparsely in the target template space.
The authors of \cite{DBLP:journals/chinaf/LiuS12} combines the minimal operational tracking results for RGB and thermal mode calculations of sparse representation coefficients. 
A weighted sparse representation regularized graph \cite{DBLP:conf/mm/LiZLZT17} is proposed to learn a robust object representation. Besides, some methods based on deep learning \cite{DBLP:journals/tip/ZhangL0Y16} and correlation filter \cite{DBLP:journals/ijon/ZhaiSLW19} have been proposed.

\vspace{-2mm}
\subsection{Image-based Training Strategy}

\noindent\textbf{Intra-image Training Strategy.} 
Reasonable use of all instances on a single image can improve the performance of the tracker. SPM \cite{DBLP:conf/cvpr/WangLXZ19} proposes a Series-Parallel Matching framework for stronger discriminative power and robustness. During the Coarse Match stage, it is expected that the target object can be discovered even though it is undergoing tremendous change in representation. GlobalTrack \cite{DBLP:conf/aaai/HuangZH20} proposes cross-query loss to enhance the discriminative power of tracker. Notebly, this training strategy is efficient because features are shared in the feature extraction part. But it is essential that the training images belong to the object detection dataset. Only when the training set is such a dataset there will be multiple target bounding box labels on a single image. 

In contrast, DFSC can be applied to the single object tracking dataset, which focuses on the cross sequence images. Different query-search image pairs are used to improve the robustness of the tracker in the CSM stage.

\begin{figure}[t]
 \begin{center}
 \includegraphics[width=1\linewidth]{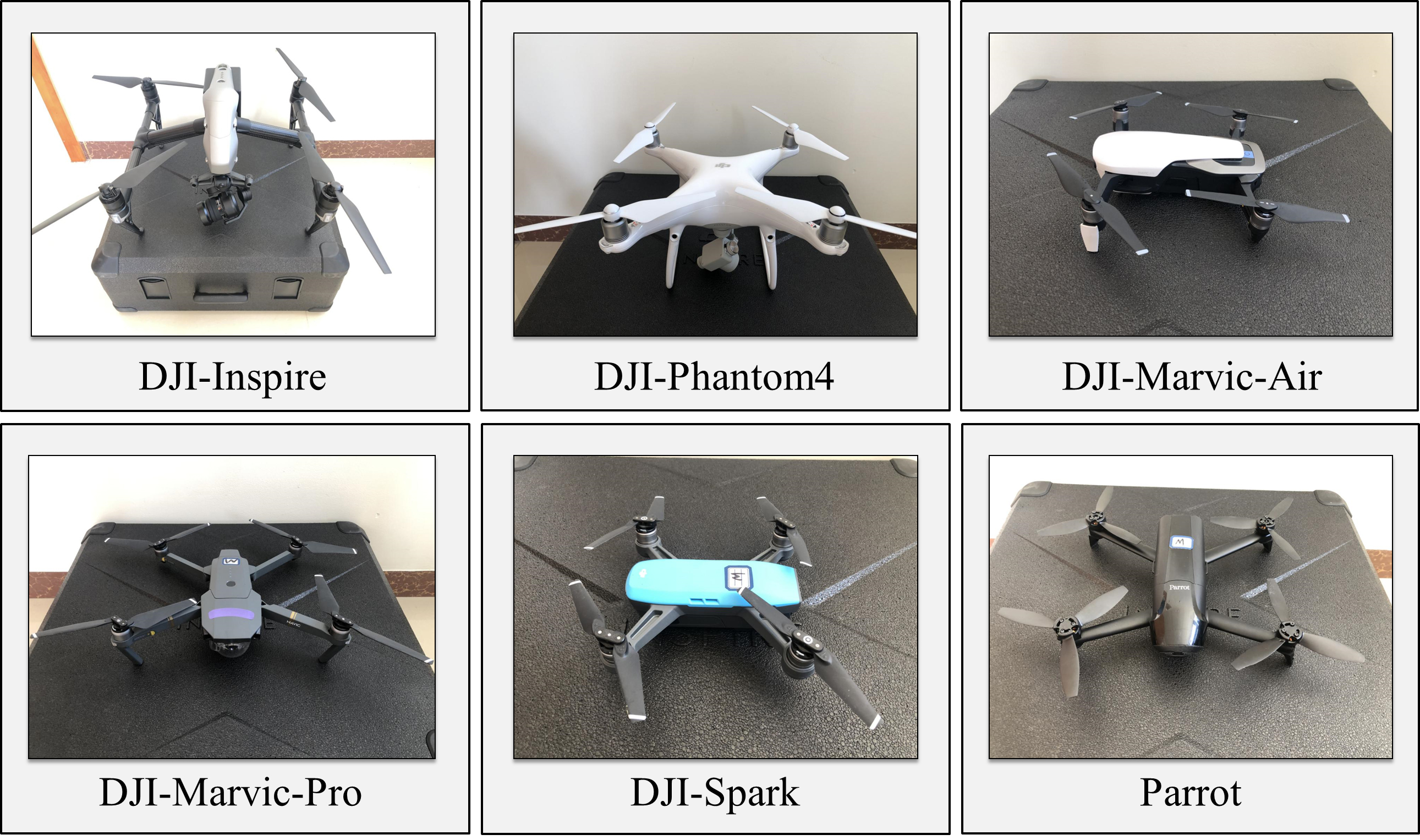}
 \end{center}
\vspace{-3mm}
\caption{
Overview of UAVs for capturing multi-modal data. 
}
\label{fig:UAV}
\vspace{-6mm}
\end{figure}

\begin{figure*}[t]
\begin{center}

	\begin{subfigure}{.66\textwidth}
		\centering
		\includegraphics[width=\textwidth]{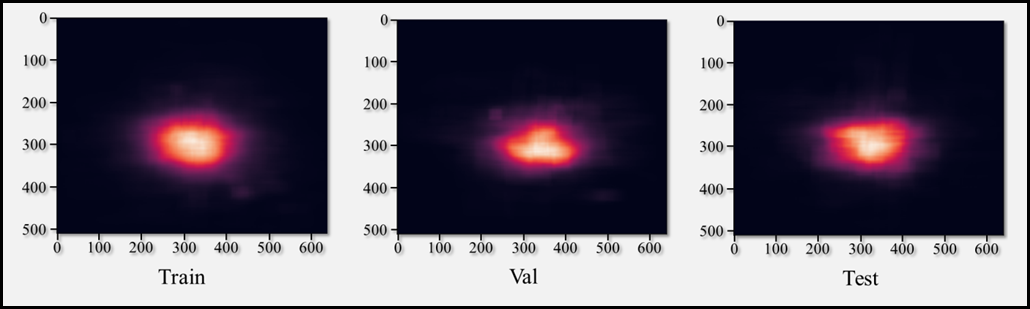}
		\caption{\label{fig:centera}Position Distribution}
	\end{subfigure}
	\begin{subfigure}{.33\textwidth}
		\centering
		\includegraphics[width=\textwidth]{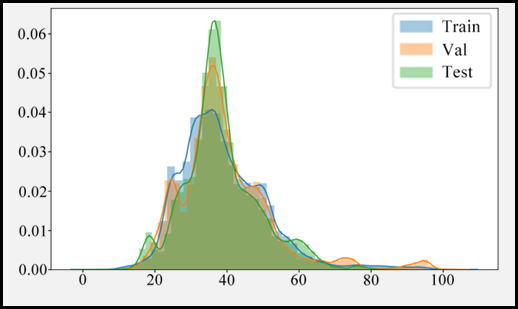}
		\caption{\label{fig:centerb}Scale Distribution}
	\end{subfigure}
\end{center}
\vspace{-4mm}
 \caption{UAV's position distribution and scale distribution in each part of Anti-UAV.
}
\label{fig:long} 
\label{fig:center}
\end{figure*}

\noindent\textbf{Inter-image Training Strategy.} 
Even if the intra-image training strategy has achieved some achievements, some works begin to focus on cross-image instances to break through current limitations. Copying and pasting over parts of an image or the whole image has been successfully utilized as a training strategy in image classification, object detection, and so on. Due to difficult cases such as occlusion in 2D human keypoints dataset, the authors \cite{DBLP:conf/eccv/KeCQL18} proposes a data augmentation method including artificially creating occlusion and constructing ambiguous image. This method \cite{DBLP:journals/corr/abs-1809-04987} is also validated on 3D human keypoint dataset, which achieves the state-of-the-art performance. In the classification task, Mixup \cite{DBLP:conf/iclr/ZhangCDL18} mixes two random samples in proportion. The results of classification are allocated in proportion. While Cutmix \cite{DBLP:conf/iccv/YunHCOYC19} is to cut out a part of the region and randomly fills the region with the pixel values of other data in the training set. The classification results are allocated according to a certain proportion. Unlike the methods mentioned above, the method in this paper copy and paste on the feature level.

Besides ``copying and pasting'' strategies, The authors of \cite{DBLP:conf/eccv/ZhuWLWYH18} introduce the existing detection datasets to enrich the positive sample data and hard negative sample data. As UAV is the only tracking category in Anti-UAV, cross-sequence pairs are used to improve the robustness of the tracker in CSM, which does not introduce additional detection datasets. At the same time, it only operates in the current batch and shares the computation at the feature extraction stage.
\vspace{-1mm}

\section{Anti-UAV Benchmark}

\subsection{Data Collection}
For the feasibility of algorithms' performance evaluation on large-scale Anti-UAV, 318 RGB-T video pairs are collected, each containing an RGB video and a thermal video.

We record various videos of several UAV types flying in the air. 
In  order  to  ensure  the  diversity  of  data, UAVs,  mainly from DJI and Parrot, are utilized to collect tracking data as shown in Fig. \ref{fig:UAV}.
The videos recorded include two lighting conditions (day and night), two light modes (infrared and visible) and diverse backgrounds (buildings, cloud, trees, \emph{etc.}). Each video is stored in an MP4 file with a frame rate of 25 FPS.

\vspace{-4mm}
\subsection{Annotation}
To guarantee the annotation quality of the Anti-UAV dataset, a progressive strategy is adopted to accurately annotate the bounding box of UAV from coarse to fine. There are three stages in the process of data annotation.

\noindent\textbf{Coarse Annotation.} In the first stage, the data is coarsely annotated:
(i) We annotate the attributes/scenes of each video, such as the size of UAVs, the condition of occlusions, possible false target, \emph{etc.}.
(ii) Each video is annotated every 25 frames. The annotation rules of this stage are as follows: If the target appears in the current frame, the flag is set as ``1" and vice versa; the UAV is labeled with an approximate rectangle. In the annotation files, the keyword \emph{``exist"} can be ``1" (True) or ``0" (False), and the keyword \emph{``get\_rect"} can be expressed as ``[x1, y1, x2, y2]" (the coordinates of the upper left and lower right corner of the bounding box).

\noindent\textbf{Fine Annotation.} According to the complexity, the videos are ranked in each scene. Then the top 10 are chosen for further annotation. After selection, there are 30 infrared and visible video pairs left. Based on the results of coarse annotation, each video frame is annotated in detail. 

\noindent\textbf{Inspection and Correction.} 
After the second stage, there are still some annotating errors such as bounding boxes with the over-large size and mislabeling of the frames with no target or a large occlusion area as ``1" , \emph{etc.}. Besides, visible videos taken at night may contain extremely blurred and distorted frames due to the pan-tilt's fast-moving. In above cases, the annotations need to be refined. At last, the annotated videos are divided into sequences every 1000 frames.

\vspace{-4mm}
\subsection{Dataset Details}

\noindent\textbf{Dataset Splitting.} Anti-UAV is divided into training set, validation set and test set. The training set and the validation set come from non-overlapped clips of the same video, but the test set is entirely independent. Besides, the test set is much more complicated than the validation set. In total 318 video pairs, 160 are divided into training set, 91 are assigned to test set, and the rest are used as validation set.

\setlength{\tabcolsep}{3.2pt}
\begin{table}
\begin{center}
\caption{Illustration of attribute annotation in Anti-UAV.}
\begin{tabular} {cp{7.3cm}}
\hline\Xhline{1.2pt}
 Attribute & Description \\
\hline\Xhline{1.2pt}
OV & Out-of-View: the target leaves the view.\\
OC & Occlusion: the target is partially occluded or heavily occluded.\\
FM & Fast Motion: the ground-truth's motion between two adjacent frames is larger than 60 pixels. \\
SV & Scale Variation: the ratio of the bounding boxes of the first frame and the current frame is out of the range [0.66, 1.5].\\
LI & Low Illumination: the illumination in the target region is low. \\
TC & Thermal Crossover: the target has a similar temperature with other objects or background surroundings.\\
LR & Low Resolution: the number of pixels inside the ground-truth bounding box is less than 400 pixels.\\\hline
\Xhline{1.2pt}
\label{tab:attr}
\end{tabular}
\end{center}
\vspace{-2.5em}
\end{table}

\begin{figure*}[t]
 \begin{center}
 \includegraphics[width=0.95\linewidth]{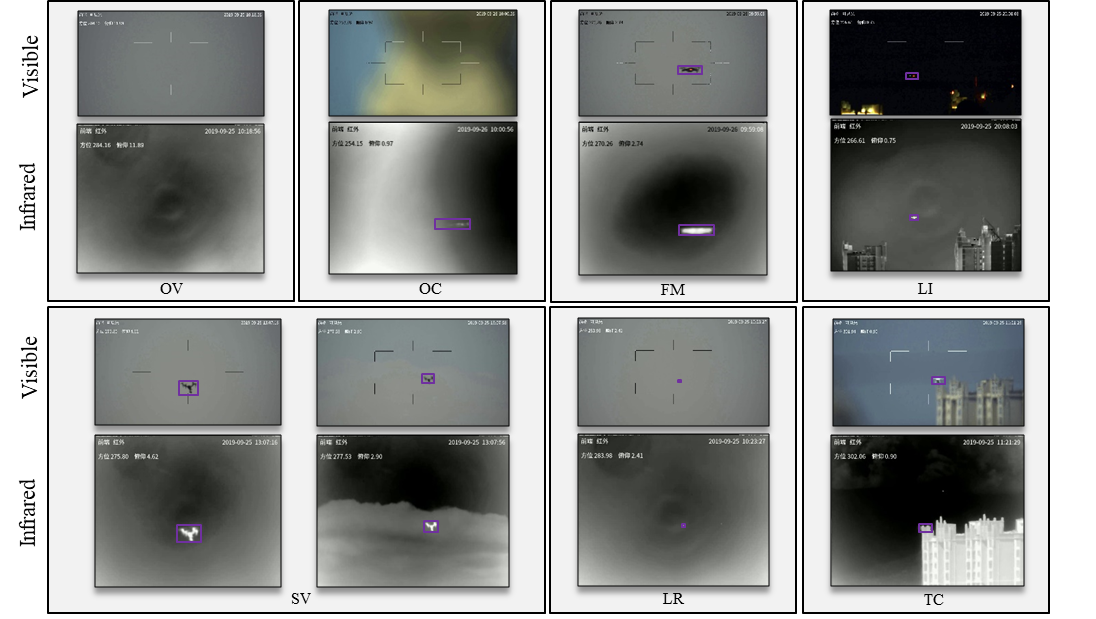}
 \end{center}
 \vspace{-2mm}
\caption{
Screenshots taken from Anti-UAV. The challenging attributes are helpful to analyze the shortcomings and advantages of trackers from all aspects. It is worth mentioning that there is no alignment between multi-modal data.
}
\vspace{-2mm}
\label{fig:example}
\end{figure*}

\begin{figure}[t]
\begin{center}
 \includegraphics[width=0.9\linewidth]{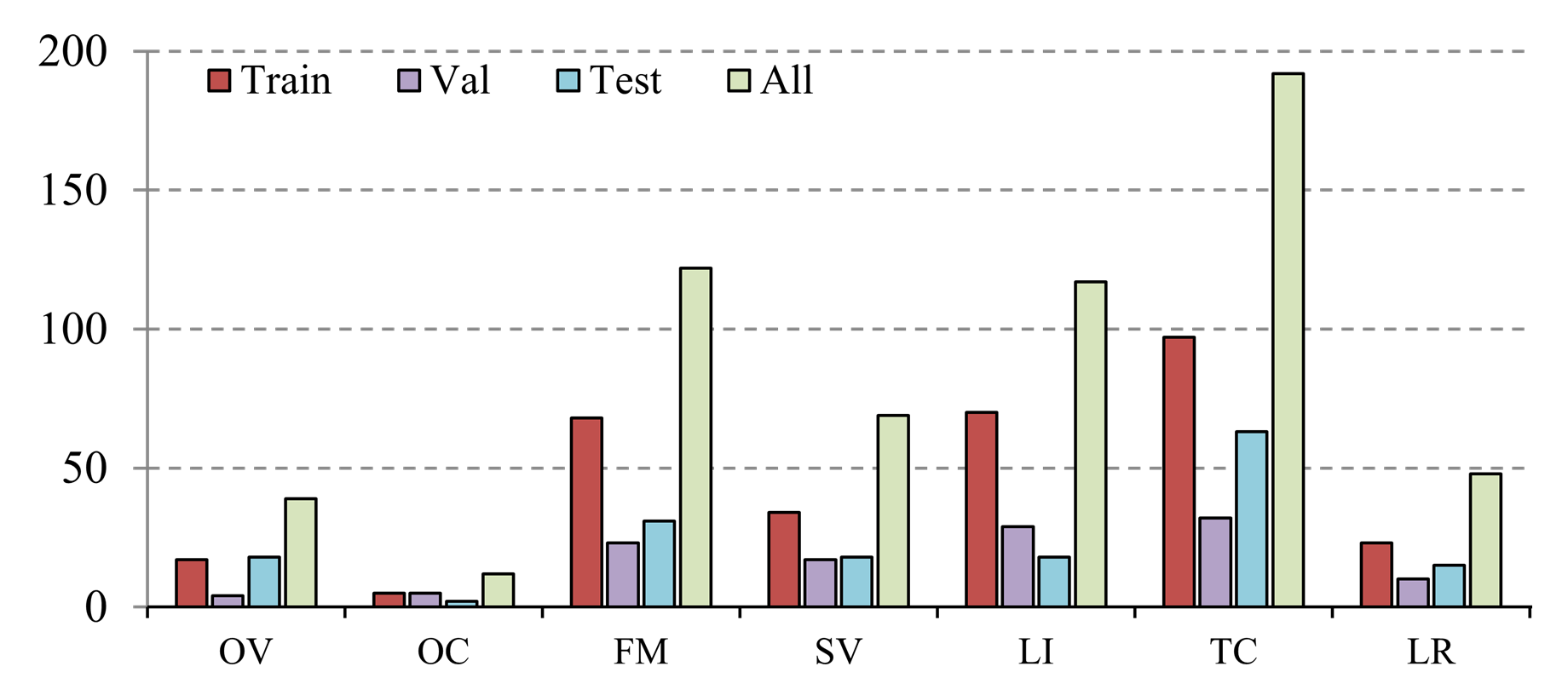}
\end{center}
\vspace{-2mm}
 \caption{The quantity of sequences with different attributes.
 }
\label{fig:statistic}
\vspace{-3mm}
\end{figure}

\noindent\textbf{Position Distribution.} As shown in the Fig. \ref{fig:centera}, the positions of the boxes are mostly concentrated in the central area of the picture, and the larger transverse variance means that the horizontal movements of UAV are in the majority. Besides, the range of motion of the target in the test set is more diverse. The test set's fluctuation is larger than that of the training set in the corresponding directions.

\noindent\textbf{Scale Distribution.} The target size of UAV fluctuates widely in the whole Anti-UAV dataset. The size of the UAV object can be calculated as $s(w, h)=\sqrt{w \times h}$. Hence, the scale distribution of UAV is given in the Fig. \ref{fig:centerb} for better analysis. It shows that the scale distributions of the three sets are relatively similar. To a certain extent, the scale distribution of the test set looks more centralized and sharper. The average values of the target size of three sets are all less than 40 pixels. 

\vspace{-3mm}
\subsection{Attributes}
The overall performance on Anti-UAV can not reflect the detailed difference between trackers. 
Provided by binary attribute annotations, the evaluation metric helps identify the pros and cons of Anti-UAV trackers. Each infrared video is annotated with the list of properties defined in Tab. \ref{tab:attr}. 

Seven attributes are provided as shown in Fig. \ref{fig:example}. For better presenting the Anti-UAV, statistics are done for the specific situation of each attribute as shown in Fig. \ref{fig:statistic}. According to the statistical results, the following conclusions can be reached:
\begin{itemize}
 \item The proportion of OV is relatively larger in the test set. However, it is not high on the whole.

 \item The case that the target among adjacent frames moves more than 60 pixels takes up a large proportion in Anti-UAV, which is one of difficulties in UAV tracking.

 \item If the SV standard of OTB is followed, almost no sequence will be considered as SV. 
 Therefore, [0.66, 1.5] may be more suitable for UAV tracking.

 \item There are more video sequences captured in the day.

 \item A large proportion of videos have the attribute of TC.
\end{itemize}

To comprehensively analyze, 
TC is divided into three categories more finely. 
According to the performance of SiamRPN++LT \cite{DBLP:conf/cvpr/LiWWZXY19} on the infrared set, the complexity levels are divided into easy, medium and hard.

\vspace{-3mm}
\subsection{Evaluation Metrics}
Anti-UAV is annotated with bounding boxes, attributes and existing flags.
Moreover, an empty bounding box list denotes a "not exist" flag. Trackers need to obtain 
the perception of UAV status. In this case, the presence of UAV in the visual range is introduced into the evaluation metric:
\begin{small}\begin{equation}\begin{aligned}\label{IOU}
SA = \sum_{t}\frac{IOU_t\times \delta(v_t > 0) + p_t \times(1 - \delta(v_t > 0))}{T}.
\end{aligned}\end{equation}\end{small}The $IoU_t$ is Intersection over Union (IoU) between each tracking bounding box and corresponding ground-truth. The $v$ are the ground-truth visibility flags (the tracker's predicted $p$ are used to measure the state accuracy). The state accuracy $SA$ is averaged over all frames in a sequence. The average state accuracy of all video sequences $mSA$ is the final evaluation result.
Precision and success are introduced to measure the performance of trackers, which is the same as \cite{DBLP:conf/cvpr/WuLY13}.

\noindent\textbf{Protocol I.} Visible or infrared video sequences are used to evaluate UAV trackers' performance, respectively. Researchers can use any training set except for those with UAV objects. It aims to verify the trackers' performance for UAV tracking without the training of the UAV dataset. Moreover, it can demonstrate the generalization of trackers in Protocol I. 

\noindent\textbf{Protocol II.} Researchers can use visible or infrared training video sequences of Anti-UAV to finetune their trackers or train them from scratch. Protocol II aims to provide a unique UAV tracking evaluation of trackers. Because there are several options for researchers to choose from, so it is essential to point out which performance comes from training data settings. Otherwise, it will bring an unfair performance comparison.

\noindent\textbf{Protocol III.} Researchers are encouraged to explore how to make the most of multi-modal data. 
The final performance is evaluated on both annotations of visible and infrared video sequences. Unlike the standard multi-modal tracking datasets published, the multi-modal data of Anti-UAV are not aligned, which is a new direction to explore using multi-modal data without alignment to track UAVs in the future.

\begin{figure*}[!htp]
 \begin{center} 
 \includegraphics[width=1\linewidth]{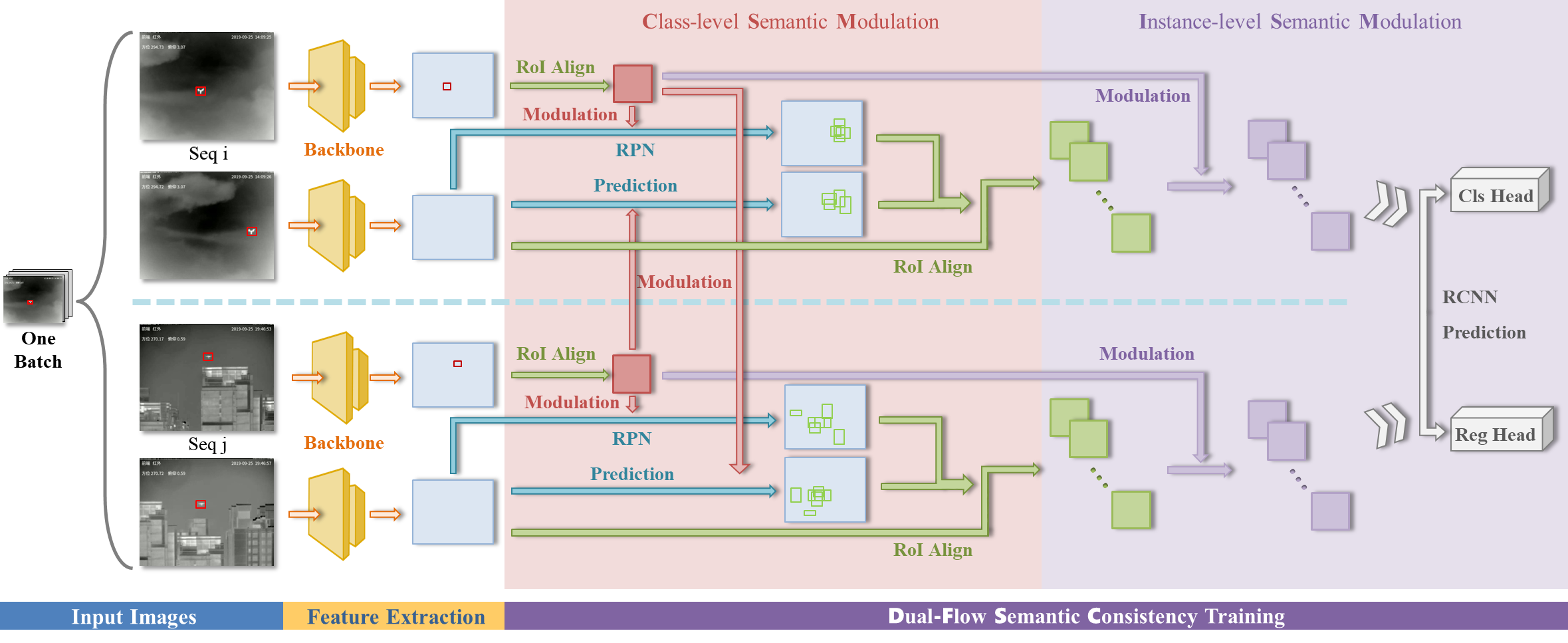}
 \end{center}
\vspace{-3mm}
\caption{
The pipeline of proposed DFSC training strategy. This figure shows the case when two video sequences (denoted as Seq i and Seq j) in a batch. There are CSM and ISM branches in the whole DFSC training process. To reduce the intraclass differences, cross-sequence UAV features are used as modulation factors to maintain class-level semantic information consistency in the CSM stage. During the ISM stage, the modulation of the query feature of the same sequence further boosts the tracker's discrimination power by reinforcing instance-level semantic consistency. 
Best viewed in color.
}
\label{fig:strategy}
\vspace{-3mm}
\end{figure*}

\vspace{-2.5mm}
\section{Methodology}
\noindent\textbf{Motivation.} 
There is only one class of objects in Anti-UAV dataset --- UAV. Therefore, even in different video sequences, the foreground information is related to each other. Inspired by this, the network can combine the features from different video sequences during training so that the feature learned can be more robust. Based on this idea, dual-flow semantic consistency (DFSC) training strategy is proposed.

\noindent\textbf{Overview.} As shown in Fig. \ref{fig:strategy}, DFSC training strategy is composed of CSM stage and ISM stage. During the CSM stage, the feature map of search image is modulated by ROI features from different video sequences' query images. In this case, the tracker will focus on potential representation of instances in entire UAV category. 
Then in ISM stage, the selected proposals will be only modulated by the ROI feature of the current video sequence's query image. In the following, the DFSC training strategy is given in detail.

\begin{figure*}[t]
 \begin{center}
 \includegraphics[width=1\linewidth, height=0.48\linewidth]{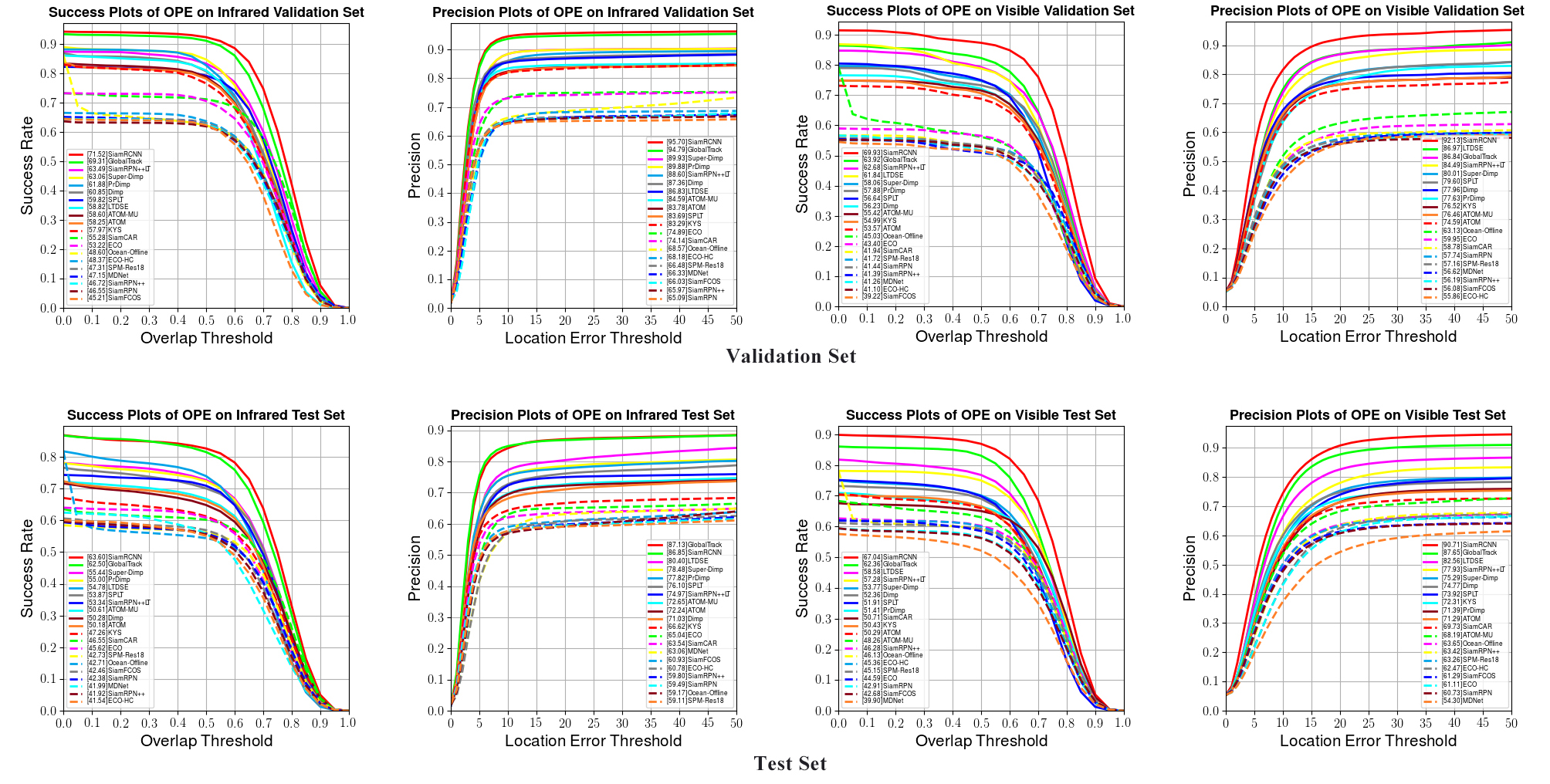}
 \end{center}
\vspace{-3mm}
\caption{The success plot and precision plot on Anti-UAV under protocol I.
For clarity, only the top 20 trackers are shown. 
}
\label{fig:p}
\end{figure*}

\setlength{\tabcolsep}{7.5pt}\begin{table*}[t] \centering 
\caption{The whole attribute-based performances $mSA$ (\%) of baseline trackers on Anti-UAV \textbf{validation} set using the evaluation protocol I. The trackers are ranked in the same way as in Tab. \ref{tab:test}. A larger number means better performance. 
The \textcolor{red}{first}-, \textcolor{blue}{second}- and \textcolor[rgb]{0,0.7,0}{third}-place trackers are labeled with red, blue and green colors respectively. Best viewed in color.
}
\begin{tabular}{l||c|c|c|c|c|c|c|c|c|c|c|c} \hline
\Xhline{1.2pt}
\multirow{3}{*}{\textbf{Tracker}} & \multicolumn{11}{c|}{\textbf{Infrared}} & \textbf{Visible} \\
\cline{2-13} &\multirow{2}{*}{OV} & \multirow{2}{*}{OC} & \multirow{2}{*}{FM} & \multirow{2}{*}{SV} & \multirow{2}{*}{LI} & \multicolumn{4}{c|}{TC} & \multirow{2}{*}{LR} & \multirow{2}{*}{All}& \multirow{2}{*}{All} \\ 
\cline{7-10} &&&&&&TC$_{easy}$&TC$_{med}$&TC$_{hard}$&TC$_{all}$&&&\\\hline \Xhline{1.2pt}
\rowcolor{mygray}MOSSE \cite{DBLP:conf/cvpr/BolmeBDL10}	&23.29	&4.63	&5.26	&8.08	&28.64	&48.70	&27.19	&8.17	&34.45	&6.20	&29.40	&21.06	\\
DAT \cite{DBLP:conf/cvpr/PosseggerMB15}	&15.98	&6.77	&8.30	&9.63	&30.76	&36.06	&20.62	&1.49	&24.64	&3.00	&29.43	&38.27	\\
\rowcolor{mygray}CSK \cite{DBLP:conf/eccv/HenriquesCMB12} &21.33	&8.90	&6.02	&8.06	&40.32	&53.32	&44.70	&7.51	&41.14	&1.96	&37.67	&35.89	\\
Staple-CA \cite{DBLP:conf/cvpr/MuellerSG17} &21.33	&10.65	&7.09	&14.09	&36.96	&55.63	&46.81	&14.54	&44.43	&10.48	&37.51	&37.23	\\
\rowcolor{mygray}MCCTH \cite{DBLP:conf/cvpr/WangZ0H0L18}	&26.51	&20.61	&8.65	&17.80	&39.75	&54.28	&42.12	&8.81	&41.29	&10.29	&38.10	&39.49	\\
Staple \cite{DBLP:conf/cvpr/BertinettoVGMT16}	&33.92	&13.82	&7.54	&17.64	&36.20	&54.80	&40.54	&13.01	&42.09	&9.81	&37.62	&38.15	\\
\rowcolor{mygray}CN \cite{DBLP:conf/cvpr/DanelljanKFW14}	&24.74	&10.12	&6.93	&14.74	&40.78	&58.34	&46.32	&11.22	&45.03	&10.25	&39.82	&36.04	\\
DCF \cite{DBLP:journals/pami/HenriquesC0B15}	&21.15	&6.93	&6.95	&13.32	&37.57	&53.64	&42.68	&9.02	&41.14	&11.44	&36.65	&36.48	\\
\rowcolor{mygray}KCF \cite{DBLP:journals/pami/HenriquesC0B15}	&22.04	&8.85	&7.50	&14.21	&37.91	&53.41	&42.94	&9.67	&41.23	&11.92	&36.82	&38.11	\\
STRCF \cite{DBLP:conf/cvpr/LiTZ0018}	&39.25	&25.34	&21.56	&27.04	&48.48	&47.92	&41.02	&4.47	&36.69	&18.34	&41.65	&44.92	\\
\rowcolor{mygray}LDES \cite{DBLP:conf/aaai/LiZHSWL19} &21.76	&8.91	&19.78	&13.55	&46.63	&54.01	&42.90	&5.05	&40.52	&8.92	&41.41	&48.98	\\
DSST \cite{DBLP:conf/bmvc/DanelljanHKF14}	&23.66	&10.46	&7.05	&14.03	&43.25	&55.31	&45.76	&13.55	&43.79	&9.29	&40.54	&37.48	\\
\rowcolor{mygray}CSRDCF \cite{DBLP:conf/cvpr/LukezicVZMK17}	&40.46	&21.13	&25.45	&27.84	&50.87	&58.66	&52.17	&14.53	&47.39	&24.77	&47.73	&41.54	\\
BACF \cite{DBLP:conf/iccv/GaloogahiFL17}&19.51	&9.00	&18.21	&20.53	&44.72	&56.47	&44.43	&7.06	&42.65	&21.99	&43.16	&43.87	\\
\rowcolor{mygray}SiamFC \cite{DBLP:conf/eccv/BertinettoVHVT16}	&42.10	&27.99	&30.28	&22.83	&53.41	&66.36	&45.80	&10.45	&48.99	&16.59	&49.34	&44.08	\\
Ocean-Online \cite{DBLP:conf/eccv/ZhangPFLH20}	&16.21	&21.02	&19.65	&24.11	&45.77	&52.66	&39.21	&6.09	&39.11	&20.33	&41.56	&46.45	\\
\rowcolor{mygray}MKCFup \cite{DBLP:conf/cvpr/TangYZW18}	&44.60	&20.13	&8.64	&22.57	&40.81	&56.95	&43.91	&13.77	&44.24	&10.25	&41.31	&40.21	\\
SiamMask \cite{DBLP:conf/cvpr/Wang0BHT19}	&40.77	&27.28	&24.22	&24.16	&47.10	&54.82	&38.76	&12.50	&41.54	&14.62	&44.34	&44.26	\\
\rowcolor{mygray}SiamDW \cite{DBLP:conf/cvpr/ZhangP19}	&43.93	&39.66	&29.40	&35.14	&56.84	&59.89	&42.31	&8.29	&44.21	&27.87	&49.46	&44.90	\\
SiamBAN \cite{DBLP:conf/cvpr/ChenZLZJ20}&20.02	&31.73	&19.03	&24.84	&48.75	&57.63	&49.44	&6.44	&44.39	&13.38	&43.60	&39.90	\\
\rowcolor{mygray}RT-MDNet \cite{DBLP:conf/eccv/JungSBH18}	&43.88	&20.82	&18.60	&27.35	&46.97	&64.97	&44.42	&13.15	&48.50	&13.52	&45.99	&44.93	\\
SPM-AlexNet \cite{DBLP:conf/cvpr/WangLXZ19}	&39.84	&24.75	&30.35	&27.70	&51.15	&56.73	&34.98	&8.20	&40.68	&14.41	&46.71	&46.51	\\
\rowcolor{mygray}ECO-HC \cite{DBLP:conf/cvpr/DanelljanBKF17} &23.16	&14.96	&25.38	&28.66	&52.66	&60.61	&48.51	&16.63	&47.96	&30.29	&49.26	&43.67	\\
Ocean-Offline \cite{DBLP:conf/eccv/ZhangPFLH20}	&26.63	&42.86	&32.09	&24.75	&55.05	&59.61	&41.66	&20.76	&46.62	&18.60	&48.74	&45.74	\\
\rowcolor{mygray}MDNet \cite{DBLP:conf/cvpr/NamH16}	&45.88	&24.16	&23.47	&31.13	&50.42	&65.96	&48.59	&14.84	&50.43	&24.85	&49.49	&45.17	\\
SiamRPN++ \cite{DBLP:conf/cvpr/LiWWZXY19} &35.32	&41.11	&28.17	&28.39	&55.85	&57.95	&44.55	&6.94	&43.44	&19.71	&48.60	&46.12	\\
\rowcolor{mygray}SiamRPN \cite{DBLP:conf/cvpr/LiYWZH18}	&32.07	&31.37	&25.73	&30.56	&52.57	&65.60	&45.24	&10.83	&48.53	&21.97	&48.16	&46.63	\\
SPM-Res18 \cite{DBLP:conf/cvpr/WangLXZ19}	&39.75	&30.78	&32.25	&29.56	&54.90	&59.16	&42.55	&12.45	&44.79	&17.95	&49.56	&46.09	\\
\rowcolor{mygray}SiamFCOS \cite{DBLP:conf/iccvw/KristanBZRGBDDN19} &41.97	&42.43	&29.01	&34.66	&53.09	&58.24	&37.02	&10.07	&42.40	&21.94	&47.71	&44.28	\\
ECO \cite{DBLP:conf/cvpr/DanelljanBKF17} &31.87	&32.03	&38.43	&38.47	&55.77	&69.24	&45.90	&21.67	&53.00	&34.25	&54.44	&46.31	\\
\rowcolor{mygray}SiamCAR \cite{DBLP:conf/cvpr/GuoWC0C20}	&25.05	&46.12	&40.84	&29.96	&63.19	&68.27	&55.99	&17.55	&54.11	&21.67	&56.70	&46.52	\\
KYS \cite{DBLP:conf/eccv/BhatDGT20}	&36.07	&55.52	&57.64	&52.68	&64.70	&66.30	&35.35	&32.05	&51.07	&48.00	&60.50	&59.79	\\
\rowcolor{mygray}ATOM \cite{DBLP:conf/cvpr/DanelljanBKF19}	&63.01	&55.82	&56.04	&46.91	&65.05	&65.36	&50.03	&25.73	&52.86	&38.53	&60.87	&58.79	\\
Dimp \cite{DBLP:conf/iccv/BhatDGT19}	&41.96	&59.85	&59.95	&55.78	&66.33	&70.19	&51.62	&30.16	&56.79	&47.52	&63.51	&61.54	\\
\rowcolor{mygray}ATOM-MU \cite{DBLP:conf/cvpr/DaiZWLLY20}	&42.73	&55.51	&58.74	&47.16	&64.83	&68.58	&46.61	&26.26	&53.83	&39.18	&61.27	&60.45	\\
SiamRPN++LT \cite{DBLP:conf/cvpr/LiWWZXY19}	&50.02	&\textcolor[rgb]{0,0.7,0}{\textbf{68.16}}	&63.39	&54.06	&\textcolor[rgb]{0,0.7,0}{\textbf{70.25}}	&\textcolor{blue}{\textbf{76.71}}	&\textcolor[rgb]{0,0.7,0}{\textbf{61.76}}	&19.25	&\textcolor[rgb]{0,0.7,0}{\textbf{60.40}}	&42.66	&\textcolor[rgb]{0,0.7,0}{\textbf{65.84}}	&\textcolor[rgb]{0,0.7,0}{\textbf{67.15}}	\\
\rowcolor{mygray}SPLT \cite{DBLP:conf/iccv/YanZWLY19}&33.39	&58.42	&55.22	&42.49	&63.18	&73.09	&58.73	&19.30	&57.73	&45.46	&60.73	&57.32	\\
PrDimp \cite{DBLP:conf/cvpr/DanelljanGT20}	&\textcolor[rgb]{0,0.7,0}{\textbf{65.22}}	&63.89	&62.85	&55.59	&66.95	&69.47	&55.79	&29.04	&57.21	&51.12	&64.54	&62.95	\\
\rowcolor{mygray}LTDSE \cite{DBLP:conf/iccvw/KristanBZRGBDDN19}	&64.39	&50.17	&59.04	&48.67	&62.69	&67.18	&55.89	&27.43	&55.66	&42.54	&61.27	&66.64	\\
Super-Dimp \footnotemark[1]	&52.35	&68.07	&\textcolor[rgb]{0,0.7,0}{\textbf{65.30}}	&\textcolor[rgb]{0,0.7,0}{\textbf{64.80}}	&67.93	&68.87	&59.85	&\textcolor[rgb]{0,0.7,0}{\textbf{34.22}}	&59.03	&\textcolor{blue}{\textbf{61.45}}	&65.76	&63.05	\\
\rowcolor{mygray} GlobalTrack \cite{DBLP:conf/aaai/HuangZH20} &\textcolor{blue}{\textbf{69.21}} &\textcolor{red}{\textbf{78.62}} &\textcolor{blue}{\textbf{73.35}} &\textcolor{blue}{\textbf{66.11}} &\textcolor{red}{\textbf{76.33}} & \textcolor[rgb]{0,0.7,0}{\textbf{76.47}} & \textcolor{blue}{\textbf{63.08}} & \textcolor{blue}{\textbf{43.45}} & \textcolor{blue}{\textbf{65.90}} &\textcolor[rgb]{0,0.7,0}{\textbf{60.26}} & \textcolor{blue}{\textbf{72.00}}& \textcolor{blue}{\textbf{67.28}}\\
SiamRCNN \cite{DBLP:conf/cvpr/VoigtlaenderLTL20} &\textcolor{red}{\textbf{73.46}}	&\textcolor{blue}{\textbf{78.24}}	&\textcolor{red}{\textbf{73.98}}	&\textcolor{red}{\textbf{67.97}}	&\textcolor{blue}{\textbf{76.19}}	&\textcolor{red}{\textbf{78.21}}	&\textcolor{red}{\textbf{69.55}}	&\textcolor{red}{\textbf{55.48}}	&\textcolor{red}{\textbf{71.07}}	&\textcolor{red}{\textbf{67.93}}	&\textcolor{red}{\textbf{74.33}}	&\textcolor{red}{\textbf{74.32}}	\\
\Xhline{1.2pt}
\hline \end{tabular} \label{tab:val} 
\vspace{-2mm}
\end{table*}

\vspace{-4mm}
\subsection{Class-level Semantic Modulation}
The class-level semantic modulation stage is to find candidate boxes containing UAV category objects, which can be regarded as a UAV detection problem. 
The training strategy of the query guided region proposal network (query-guided RPN) as introduced in GlobalTrack is modified. The specific training strategy modulates the search area by using cross-sequence query. The general UAV modulation feature obtained in the CSM stage can be defined as $\hat{t}$. Specifically, the cross-sequence query modulation is defined as follows,
\begin{small}\begin{equation}\label{t}
\hat{t}_{ij} = f_{CSM}(z_{i}, x_{j}) =f_{out}((f_{z}(z_{i}) \otimes f_{x}(x_{j}))),
\end{equation}\end{small}where $z_{i}$ denotes the ROI features of the query in $i$-th sequence, $x_{j}$ denotes the feature of the search image of $j$-th sequence extracted from backbone. Specially, $f_{CSM}$ is the modulater to modulate the intra-sequence and cross-sequence using different combinations of $z_{i}$ and $x_{j}$. $\hat{t_{ij}}$ retains the size of $x_{j}$, and it denotes the modulated feature which will be utilized to generate the proposals.
 $f_{out}$ is utilized to align the feature channel number of $\hat{t_{ij}}$ and $x_{j}$, $f_{z}$ and $f_{x}$ respectively act on $z_{i}$ and $x_{j}$ to obtain the projected feature. 	$\otimes$ represents convolution operator.

It is worth mentioning that the value range of $i$ and $j$ is between 0 and the current batch size $n$. When $i$ and $j$ are not equal, the cross sequence's image modulation method will be adopted. When $i$ and $j$ are equal, the image modulation method degenerates to the intra-sequence.

According to the above definition, the classification and regression part of the tracker are trained in the CSM stage. In training, the loss function is defined as follows,
\begin{small}\begin{equation}\begin{aligned}\label{CSM}
L_{CSM}(z_i,x_i, z_j, x_j) &= \left.L_{same} + \alpha L_{cross}
\right.
\\
\phantom{=\;\;}
&= \sum_{i,j \in n, i = j }L_{rpn}(\hat{t}_{ij}) + \alpha\sum_{i,j \in n, i \neq j }L_{rpn}(\hat{t}_{ij}).
\end{aligned}\end{equation}\end{small}Here $\alpha$ is a weight coefficient for adjusting the ratio between $L_{cross}$ and $L_{same}$. Both $L_{cross}$ and $L_{same}$ are loss functions of RPN. Here ``cross" denotes RPN prediction after cross-sequence modulation while ``same" denotes PRN prediction after intra-sequence modulation. Specifically, the training loss function of RPN can be expressed as follows
\begin{small}\begin{equation}\label{rpn}
L_{rpn}(\hat{t}_{ij}) = \frac{1}{N_{cls}}\sum_{n}L_{cls}(s_{n},s_{n}^{*}) + \beta\frac{1}{N_{reg}}\sum_{n}L_{reg}(p_{n},p_{n}^{*}).
\end{equation}\end{small}Here $\beta$ is a weight utilized to balance the classification and regression losses, $s_{n}$ and $s_{n}^{*}$ respectively represent the estimated classification score and corresponding groundtruth, while $p_{n}$ and $p_{n}^{*}$ are the location of the $n$-th proposal and the corresponding groundtruth.

\vspace{-4mm}
\subsection{Instance-level Semantic Modulation}
In the previous stage, proposals similar to UAV class are chosen. While in ISM stage, the emphasis is on instance-related information. At this point, the tracker aims to distinguish instances from those with similar appearance information or from complex background.

Given the query of the sequence to which the current feature belongs, proposals will be utilized for classification and bounding box refinement. The instance-level semantic modulation between $z$ and $x_{k}$ is performed as:
\begin{small}\begin{equation}\label{t}
\hat{t}_{k} = f_{ISM}(z, x_{k}) =f_{out}^{\prime}((f_{z}^{\prime}(z) \odot f_{x}^{\prime}(x_{k}))),
\end{equation}\end{small}where $x_{k}$ represents the selected $k$-th proposals and $z$ represents the ROI feature of the query image that in the same sequence which the current feature $x_{k}$ comes from. $f_{ISM}$ is the modulater to modulate instance-special information into selected proposals. $f_{out}^{\prime}$ keeps $\hat{t}_{k}$ and $x_{k}$ the same size. $f_{z}^{\prime}$ and $f_{x}^{\prime}$ represent the feature projection modules for $z$ and $x_{k}$, respectively. $\odot$ denotes the Hadamard production.

Then, the GlobalTrack QG-RCNN training is proceeded. The modulated ROI feature $\hat{t_{k}}$ will be performed classification and regression to get the results of the tracker as follows,
\begin{small}\begin{equation}\label{ISM}
L_{ISM}(z,x) = \frac{1}{N_{pnum}}\sum_{k}L_{rcnn}(\hat{t}_{k}).
\end{equation}\end{small}Here $N_{pnum}$ denotes the number of the selected proposals from the CSM stage. For every modulated ROI feature $\hat{t}_{k}$, the loss function can be formulated as
\begin{small}\begin{equation}\label{rcnn}
L_{rcnn}(\hat{t}_{k}) = L_{cls}^{\prime}(s_{n}^{\prime},s_{n}^{{\prime}*}) + \beta L_{reg}^{\prime}(p_{n}^{\prime},p_{n}^{{\prime}*}),
\end{equation}\end{small}where $s_{n}^{\prime}$ and $s_{n}^{{\prime}*}$ represent the predicted confidence score and corresponding groundtruth, respectively. $p_{n}^{\prime}$ and $p_{n}^{{\prime}*}$ are the location of the $n$-th proposal and the corresponding groundtruth.

\vspace{-3mm}
\section{Experiments}
\vspace{-2mm}
\subsection{Trackers}

Deep learning based trackers \cite{DBLP:conf/eccv/BertinettoVHVT16,DBLP:conf/eccv/ZhangPFLH20,DBLP:conf/cvpr/Wang0BHT19,DBLP:conf/cvpr/ZhangP19,DBLP:conf/cvpr/ChenZLZJ20,DBLP:conf/eccv/JungSBH18,DBLP:conf/cvpr/WangLXZ19,DBLP:conf/cvpr/NamH16,DBLP:conf/cvpr/LiWWZXY19,DBLP:conf/cvpr/LiYWZH18,DBLP:conf/iccvw/KristanBZRGBDDN19,DBLP:conf/cvpr/GuoWC0C20,DBLP:conf/eccv/BhatDGT20,DBLP:conf/cvpr/DanelljanBKF19,DBLP:conf/iccv/BhatDGT19,DBLP:conf/cvpr/DaiZWLLY20,DBLP:conf/iccv/YanZWLY19,DBLP:conf/cvpr/DanelljanGT20,DBLP:conf/aaai/HuangZH20,DBLP:conf/cvpr/VoigtlaenderLTL20} and correlation filters based trackers \cite{DBLP:conf/cvpr/BolmeBDL10,DBLP:conf/cvpr/PosseggerMB15,DBLP:conf/eccv/HenriquesCMB12,DBLP:conf/cvpr/MuellerSG17,DBLP:conf/cvpr/WangZ0H0L18,DBLP:conf/cvpr/BertinettoVGMT16,DBLP:journals/pami/HenriquesC0B15,DBLP:conf/cvpr/DanelljanKFW14,DBLP:conf/cvpr/LiTZ0018,DBLP:conf/aaai/LiZHSWL19,DBLP:conf/bmvc/DanelljanHKF14,DBLP:conf/cvpr/LukezicVZMK17,DBLP:conf/iccv/GaloogahiFL17,DBLP:conf/cvpr/TangYZW18,DBLP:conf/cvpr/DanelljanBKF17} are used for comparison under Protocol I and briefly described below. 

Trackers based on deep learning need to train on large-scale datasets. In recent years, due to the rapid development of deep learning and hardware, object tracking based on deep learning has made great progress. In this section, different kinds of trackers are introduced. Siamese network family occupies a very important position in the field of single object tracking. 

\vspace{-3mm}

\subsection{Implementation Details}
\noindent\textbf{Hyper-parameters.} For protocol II, GlobalTrack is the baseline. There are two settings of hyper-parameters for visible and infrared datasets, respectively. 

For Anti-UAV visible dataset, the pre-trained model provided by GlobalTrack is utilized as the initial model. There are 12 epochs in total, and the initial learning rate is set to 0.02, which is then set to 0.002 and 0.0002 at the 8th epoch and the 11th epoch, respectively.
Concerning Anti-UAV infrared dataset, the initial weights transferred from Faster RCNN are adopted and then trained with 18 epochs. The learning rate is also set to 0.02, and we decay it to 0.002 and 0.0002 at the 12th, 15th epoch, respectively.

In the training phase of CSM and ISM, the classification and regression losses are respectively cross-entropy and smooth l1, while the batch size is set to 2 per GPU.

\noindent\textbf{Training Set.} Anti-UAV infrared dataset is used for training infrared UAV tracker. However, in the visible UAV tracker training, the pre-trained GlobalTrack model utilizes a combination of MS COCO \cite{DBLP:conf/eccv/LinMBHPRDZ14}, LaSOT and GOT-10k. During the finetune stage, only the Anti-UAV visible dataset is adopted.

\setlength{\tabcolsep}{7.5pt}\begin{table*} \centering
\caption{The performances $mSA$ (\%) of baseline trackers on the Anti-UAV \textbf{test} set using the evaluation protocol I. The trackers are ranked by their state accuracy scores of infrared video on test set. Larger number means better performance. 
}
\begin{tabular}{l||c|c|c|c|c|c|c|c|c|c|c|c} \hline 
\Xhline{1.2pt} \multirow{3}{*}{\textbf{Tracker}} & \multicolumn{11}{c|}{\textbf{Infrared}} & \textbf{Visible} \\ 
\cline{2-13} &\multirow{2}{*}{OV} & \multirow{2}{*}{OC} & \multirow{2}{*}{FM} & \multirow{2}{*}{SV} & \multirow{2}{*}{LI} & \multicolumn{4}{c|}{TC} & \multirow{2}{*}{LR} & \multirow{2}{*}{All} & \multirow{2}{*}{All} \\ 
\cline{7-10} &&&&&&TC$_{easy}$&TC$_{med}$&TC$_{hard}$&TC$_{all}$&&&\\\hline 
\Xhline{1.2pt}
\rowcolor{mygray}MOSSE \cite{DBLP:conf/cvpr/BolmeBDL10}	&8.89	&24.16	&6.02	&4.06	&3.56	&15.23	&10.34	&5.80	&10.13	&3.80	&13.47	&15.23	\\
DAT \cite{DBLP:conf/cvpr/PosseggerMB15}	&8.01	&21.94	&5.33	&13.53	&3.11	&40.57	&13.28	&6.50	&17.41	&3.76	&22.68	&27.19	\\
\rowcolor{mygray}CSK \cite{DBLP:conf/eccv/HenriquesCMB12} &11.51	&26.97	&9.56	&12.35	&2.71	&46.51	&15.63	&5.30	&19.54	&3.29	&24.26	&28.38	\\
Staple-CA \cite{DBLP:conf/cvpr/MuellerSG17} &15.60	&41.11	&13.29	&9.03	&3.64	&46.25	&18.27	&7.38	&21.37	&5.53	&25.44	&31.40	\\
\rowcolor{mygray}MCCTH \cite{DBLP:conf/cvpr/WangZ0H0L18} &11.58	&33.21	&9.84	&9.08	&4.95	&41.09	&20.33	&6.61	&21.02	&5.13	&25.85	&29.96	\\
Staple \cite{DBLP:conf/cvpr/BertinettoVGMT16}	&14.74	&44.09	&11.56	&11.67	&3.70	&44.56	&21.51	&6.82	&22.44	&5.26	&26.50	&29.71	\\
\rowcolor{mygray}CN \cite{DBLP:conf/cvpr/DanelljanKFW14}	&14.41	&39.75	&10.66	&15.18	&3.54	&59.75	&28.02	&7.93	&29.33	&4.81	&31.72	&28.65	\\
DCF	 \cite{DBLP:journals/pami/HenriquesC0B15} &14.85	&36.89	&12.28	&11.55	&3.18	&60.32	&30.26	&8.80	&30.80	&4.25	&32.55	&38.39	\\
\rowcolor{mygray}KCF \cite{DBLP:journals/pami/HenriquesC0B15} &16.14	&37.56	&12.60	&11.66	&3.55	&60.23	&30.80	&9.17	&31.16	&4.36	&32.88	&39.40	\\
STRCF \cite{DBLP:conf/cvpr/LiTZ0018}	&15.72	&44.39	&14.69	&20.18	&7.31	&59.49	&28.26	&10.89	&30.23	&7.84	&33.77	&45.19	\\
\rowcolor{mygray}LDES \cite{DBLP:conf/aaai/LiZHSWL19}	&16.83	&40.97	&17.86	&16.33	&8.40	&60.88	&28.07	&10.46	&30.33	&7.54	&34.46	&49.13	\\
DSST \cite{DBLP:conf/bmvc/DanelljanHKF14}	&14.45	&41.19	&12.59	&15.88	&3.57	&69.31	&30.86	&9.37	&33.27	&4.85	&35.18	&35.67	\\
\rowcolor{mygray}CSRDCF \cite{DBLP:conf/cvpr/LukezicVZMK17}	&13.70	&46.26	&12.96	&19.14	&4.97	&61.55	&32.05	&10.22	&32.37	&7.26	&35.29	&47.10	\\
BACF \cite{DBLP:conf/iccv/GaloogahiFL17} &16.17	&41.91	&15.66	&16.70	&4.13	&70.16	&33.02	&8.53	&34.28	&4.91	&36.78	&47.52	\\
\rowcolor{mygray}SiamFC \cite{DBLP:conf/eccv/BertinettoVHVT16}	&18.59	&60.83	&21.46	&23.58	&13.55	&63.82	&29.00	&11.02	&31.60	&10.36	&36.97	&45.69	\\
Ocean-Online \cite{DBLP:conf/eccv/ZhangPFLH20}	&17.98	&41.56	&14.72	&17.35	&3.73	&68.51	&32.02	&9.27	&33.63	&4.45	&37.22	&48.11	\\
\rowcolor{mygray}MKCFup \cite{DBLP:conf/cvpr/TangYZW18}	&16.44	&43.35	&15.60	&14.15	&3.48	&70.74	&35.55	&8.92	&35.76	&4.88	&37.41	&39.52	\\
SiamMask \cite{DBLP:conf/cvpr/Wang0BHT19}	&29.09	&53.55	&18.73	&22.03	&8.14	&59.32	&30.42	&17.91	&33.27	&10.08	&37.44	&45.92	\\
\rowcolor{mygray}SiamDW \cite{DBLP:conf/cvpr/ZhangP19} &19.36	&38.14	&17.65	&22.18	&8.52	&57.68	&36.28	&13.73	&34.60	&9.44	&38.01	&49.86	\\
SiamBAN \cite{DBLP:conf/cvpr/ChenZLZJ20}&14.92	&33.72	&16.42	&18.84	&4.78	&72.67	&39.33	&16.90	&40.33	&6.15	&40.86	&44.71	\\
\rowcolor{mygray}RT-MDNet \cite{DBLP:conf/eccv/JungSBH18} &19.66	&50.00	&20.88	&21.38	&12.42	&65.38	&37.37	&16.21	&37.55	&7.98	&41.05	&42.59	\\
SPM-AlexNet \cite{DBLP:conf/cvpr/WangLXZ19}	&28.82	&54.65	&22.99	&21.89	&10.96	&72.10	&35.75	&16.00	&38.19	&10.86	&41.33	&54.09	\\
\rowcolor{mygray}ECO-HC	 \cite{DBLP:conf/cvpr/DanelljanBKF17} &20.48	&50.46	&20.72	&24.69	&6.74	&\textcolor[rgb]{0,0.7,0}{\textbf{77.18}}	&41.74	&14.76	&41.91	&10.21	&42.39	&48.91	\\
Ocean-Offline \cite{DBLP:conf/eccv/ZhangPFLH20} &25.11	&53.63	&23.74	&21.07	&12.22	&71.65	&40.67	&9.26	&38.58	&10.74	&42.51	&47.45	\\
\rowcolor{mygray}MDNet \cite{DBLP:conf/cvpr/NamH16}	&28.90	&73.29	&24.19	&20.60	&12.95	&65.92	&42.13	&15.44	&39.79	&12.14	&42.95	&43.94	\\
SiamRPN++ \cite{DBLP:conf/cvpr/LiWWZXY19} &22.96	&48.88	&20.44	&21.27	&9.63	&75.16	&40.24	&16.46	&41.21	&10.76	&43.01	&51.69	\\
\rowcolor{mygray}SiamRPN \cite{DBLP:conf/cvpr/LiYWZH18} &25.35	&46.57	&24.50	&21.90	&14.92	&74.95	&39.37	&11.54	&39.32	&11.18	&43.39	&47.96	\\
SPM-Res18 \cite{DBLP:conf/cvpr/WangLXZ19} &27.02	&55.47	&23.73	&26.13	&10.85	&76.39	&40.14	&14.16	&40.77	&12.92	&44.06	&50.42	\\
\rowcolor{mygray}SiamFCOS \cite{DBLP:conf/iccvw/KristanBZRGBDDN19}&28.74	&53.54	&23.39	&26.24	&10.81	&73.10	&42.28	&17.89	&42.16	&12.40	&44.37	&48.54	\\
ECO	\cite{DBLP:conf/cvpr/DanelljanBKF17} &24.38	&45.92	&23.90	&23.36	&11.38	&75.76	&48.21	&14.72	&44.76	&7.97	&46.51	&47.31	\\
\rowcolor{mygray}SiamCAR \cite{DBLP:conf/cvpr/GuoWC0C20} &28.90	&48.06	&29.03	&27.63	&17.82	&\textcolor{red}{\textbf{78.88}}	&45.27	&13.01	&43.52	&12.52	&47.82	&54.79	\\
KYS	\cite{DBLP:conf/eccv/BhatDGT20} &40.80	&55.25	&35.23	&35.70	&33.30	&73.71	&46.77	&17.58	&44.42	&24.61	&49.32	&55.85	\\
\rowcolor{mygray}ATOM \cite{DBLP:conf/cvpr/DanelljanBKF19} &40.17	&53.91	&36.45	&34.39	&36.70	&73.24	&54.37	&23.58	&49.77	&25.84	&52.19	&55.68	\\
Dimp \cite{DBLP:conf/iccv/BhatDGT19} &40.87	&55.29	&36.57	&40.03	&32.42	&73.53	&48.82	&29.93	&48.91	&25.57	&52.47	&58.25	\\
\rowcolor{mygray}ATOM-MU \cite{DBLP:conf/cvpr/DaiZWLLY20}&38.67	&53.84	&35.37	&35.73	&35.44	&74.30	&52.16	&26.81	&49.84	&24.85	&52.61	&54.02	\\
SiamRPN++LT	\cite{DBLP:conf/cvpr/LiWWZXY19} &45.50	&71.71	&47.09	&43.61	&44.70	&75.88	&52.75	&18.66	&48.15	&32.35	&54.34	&61.17	\\
\rowcolor{mygray}SPLT \cite{DBLP:conf/iccv/YanZWLY19}&49.92	&51.73	&\textcolor[rgb]{0,0.7,0}{\textbf{51.75}}	&41.69	&54.76	&72.46	&50.82	&26.39	&48.65	&37.89	&54.63	&53.10	\\
PrDimp \cite{DBLP:conf/cvpr/DanelljanGT20}&\textcolor[rgb]{0,0.7,0}{\textbf{57.43}}	&\textcolor{red}{\textbf{79.52}}	&49.40	&\textcolor[rgb]{0,0.7,0}{\textbf{50.05}}	&49.09	&74.29	&53.68	&31.43	&51.90	&37.42	&56.50	&57.02	\\
\rowcolor{mygray}LTDSE \cite{DBLP:conf/iccvw/KristanBZRGBDDN19}	&56.25	&75.55	&53.65	&49.89	&\textcolor[rgb]{0,0.7,0}{\textbf{56.72}}	&71.19	&52.04	&\textcolor[rgb]{0,0.7,0}{\textbf{37.79}}	&52.22	&\textcolor[rgb]{0,0.7,0}{\textbf{48.70}}	&56.51	&\textcolor[rgb]{0,0.7,0}{\textbf{64.29}}	\\
Super-Dimp \footnotemark[1] &53.37	&\textcolor[rgb]{0,0.7,0}{\textbf{78.79}}	&46.59	&47.45	&46.77& 75.39	&\textcolor[rgb]{0,0.7,0}{\textbf{56.50}}	&31.99	&\textcolor[rgb]{0,0.7,0}{\textbf{53.69}}	&35.58	&\textcolor[rgb]{0,0.7,0}{\textbf{57.72}}	&59.49	\\
\rowcolor{mygray}GlobalTrack \cite{DBLP:conf/aaai/HuangZH20} &\textcolor{red}{\textbf{68.98}} & \textcolor{blue}{\textbf{79.47}} & \textcolor{blue}{\textbf{63.42}} & \textcolor{red}{\textbf{57.34}}	 & \textcolor{blue}{\textbf{67.78}} & 74.38& \textcolor{blue}{\textbf{60.24}}& \textcolor{red}{\textbf{43.02}}& \textcolor{blue}{\textbf{58.46}} &  \textcolor{blue}{\textbf{58.48}} & \textcolor{blue}{\textbf{63.86}}& \textcolor{blue}{\textbf{66.24}}\\
SiamRCNN \cite{DBLP:conf/cvpr/VoigtlaenderLTL20} &\textcolor{blue}{\textbf{68.17}}	&78.49	&\textcolor{red}{\textbf{67.66}} &\textcolor{blue}{\textbf{57.23}}	&\textcolor{red}{\textbf{73.92}}	&\textcolor{blue}{\textbf{78.78}}	&\textcolor{red}{\textbf{61.89}}	&\textcolor{blue}{\textbf{42.48}}	&\textcolor{red}{\textbf{60.10}}	&\textcolor{red}{\textbf{64.04}}	&\textcolor{red}{\textbf{65.41}}	&\textcolor{red}{\textbf{70.83}}	\\
\Xhline{1.2pt}
\hline \end{tabular} 
\label{tab:test} 
 \end{table*}

\footnotetext[1]{https://github.com/visionml/pytracking}

\vspace{-3mm}
\subsection{Evaluations under Protocol I}
\noindent\textbf{Overall Performance.} Fig. \ref{fig:p} shows the success plot and the precision plot of one pass evaluation (OPE) on Anti-UAV. SiamRCNN achieves the best precision score of 95.70\% and a success score of 71.52\% on the infrared validation set.
On the infrared test set, SiamRCNN obtains state-of-the-art results with a 63.60\% success score; however, GlobalTrack shows top performance with precision score of 87.13\%. Moreover, SiamRCNN achieves more remarkable performance improvement than the infrared tracking sequence on the visible tracking sequence. For instance, SiamRCNN outperforms 4.68 points in success score and 3.06 points in precision score over the second tracker GlobalTrack on the visible test set. However, it is worth mentioning that the inference speed of SiamRCNN is far slower than other trackers. In the meanwhile, ECO-HC is the best correlation filter tracker on the infrared and visible test set, which is with a 60.78\% precision score and a 41.54\% success score on the infrared test set. The best deep tracker with online learning is Super-Dimp with 78.48\% precision score and 55.44\% success score.

In Tab. \ref{tab:val} and Tab. \ref{tab:test}, each tracker is used without any modification. The evaluation results in $mSA$ (\%) is given. For infrared sequence, SiamRCNN achieves the best state accuracy score of 74.33\% on the validation set and 65.41\% on the test set. According to the experimental results, the tracker based on long-term tracking is more likely to achieve higher performances. The long-term tracker's underlying assumption is that tracking the target out of view is possible. In this setting, long-term tracking will generally have a larger search area or search in the whole image to enable the tracker to obtain the vanishing target reappearing position.

Nevertheless, the above mentioned long-term tracker is challenging to achieve a real-time requirement. The order of the tracker is from low to high according to the performance on the test set. Trackers based on deep learning generally have a higher performance. In most cases, UAV tracking under visible video sequence has a better solution.

\noindent\textbf{Attribute-based Performance.} To analyze different challenges faced by existing trackers, all trackers are evaluated on seven attributes. The results of several challenging attributes are shown in Tab. \ref{tab:val} and Tab. \ref{tab:test}.

For instance, OV generally does not appear in the short-term tracker setting. 
The short-term tracking sequence is usually not as long as the sequence in the Anti-UAV dataset.
Meanwhile, because the UAV is generally far away from the camera, it is also challenging to track such a small target. Therefore, many trackers do not perform well on video sequences with LR attribute, especially correlation filter tracker, due to the trackers' anchor setting. Tracking sequences with TC attribute make up a large part of Anti-UAV. 
$\rm TC_{easy}$, $\rm TC_{med}$ and $\rm TC_{hard}$ is generated by the tracking difficulty of the corresponding infrared sequence. $\rm TC_{hard}$ is the most challenging attribute. When tracking UAVs in the video sequences with $\rm TC_{hard}$ attribute, it is not easy to distinguish the UAV from the buildings without fine-tuning using the Anti-UAV training set.

On most attributes, SiamRCNN and GlobalTrack perform much better than other trackers for their more sophisticated processing designs. These two trackers achieve comparable performance except for FM, LR and LI. The superiority that over 4.00\% $mSA$ of the above three attributes makes SiamRCNN a lead. And on OV and SV, GlobalTrack is slightly ahead while it is reverse on TC. As for the validation set, GlobalTrack has a slight lead over the second tracker on OC and LI, while the performances of SiamRCNN on other attributes are all the best. Especially on OV, TC and LR, SiamRCNN performs far better than other trackers. Thus SiamRCNN performs well on each of the seven attributes.

\setlength{\tabcolsep}{7.2pt}\begin{table*} \centering
\caption{Comparisons in terms of $mSA$ (\%) with different training methods. The definition of large-scale and normal training strategy can be seen in section \ref{sec:define}. The \textbf{Bold} indicates the best performance on corresponding dataset.}
\begin{tabular}{l||c|c|c|c|c|c|c|c|c|c|c|c|c} \hline 
\Xhline{1.2pt}
\multirow{3}{*}{\textbf{Method}} & \multirow{3}{*}{\textbf{Type}}& \multicolumn{11}{c|}{\textbf{Infrared}} & \textbf{Visible} \\ 
\cline{3-14} & &\multirow{2}{*}{OV} & \multirow{2}{*}{OC} & \multirow{2}{*}{FM} & \multirow{2}{*}{SV} & \multirow{2}{*}{LI} & \multicolumn{4}{c|}{TC} & \multirow{2}{*}{LR} & \multirow{2}{*}{All} & \multirow{2}{*}{All} \\ 
\cline{8-11} &&&&&&&TC$_{easy}$&TC$_{med}$&TC$_{hard}$&TC$_{all}$&&&\\\hline 
\Xhline{1.2pt}
\rowcolor{mygray} large-scale &\multirow{3}{*}{val}&69.21 &78.62 &73.35 &66.11 &76.33 & 76.47 & 63.08 & 43.45 & 65.90 &60.26 & 72.00& 67.28\\
normal &&\textbf{78.09} & 81.42 & 78.66 & 76.61 & 79.95 & 81.23 & 76.56 & 74.04 & 78.49 & 73.55 & 79.60 & 73.25 \\
\rowcolor{mygray}DFSC (Ours)&&77.72 & \textbf{82.70} & \textbf{79.34} & \textbf{77.58} & \textbf{80.31} & \textbf{81.48} & \textbf{76.83} & \textbf{75.65} & \textbf{79.04} & \textbf{74.33} & \textbf{80.09} & \textbf{73.73}\\\hline
\Xhline{1.2pt}
large-scale & \multirow{3}{*}{test} &68.98	 &\textbf{79.47} & \textbf{63.42} 	&\textbf{57.34}	 &\textbf{67.78} & 74.38& 60.24& 43.02& 58.46 & \textbf{58.48} & 63.86& 66.24\\
\rowcolor{mygray}normal & test &\textbf{70.44} & 68.94 & 60.66 & 55.48 & 59.78 & 77.07 & 64.64 & 44.61 & 61.68 & 52.94 & 65.36 & 69.27 \\
DFSC (Ours)&& 70.16 & 68.07 & 60.95 & 55.55 & 60.13 & \textbf{77.96} & \textbf{65.85} & \textbf{45.59} & \textbf{62.75} & 53.10 & \textbf{66.04} & \textbf{69.84} \\
\Xhline{1.2pt}
\hline \end{tabular} \label{tab:ourval} \end{table*}

\setlength{\tabcolsep}{8.5pt}\begin{table*} \centering
\caption{Ablation study in terms of $mSA$ (\%) on the \textbf{test} set.
DFSC-all represents that ISM stage uses the same class-level modulation as CSM. 
DFSC-cls and DFSC-reg denote only corresponding task is introduced in the CSM stage under the same conditions. The ratio $\alpha$ is a weight to balance $L_{cross}$ and $L_{same}$ as shown in Eq. \ref{CSM}. 
}
\begin{tabular}{l||c|c|c|c|c|c|c|c|c|c|c|c} \hline 
\Xhline{1.2pt}
\multirow{3}{*}{\textbf{Method}} & \multicolumn{11}{c|}{\textbf{Infrared}} & \textbf{Visible} \\ 
\cline{2-13} &\multirow{2}{*}{OV} & \multirow{2}{*}{OC} & \multirow{2}{*}{FM} & \multirow{2}{*}{SV} & \multirow{2}{*}{LI} & \multicolumn{4}{c|}{TC} & \multirow{2}{*}{LR} & \multirow{2}{*}{All} & \multirow{2}{*}{All} \\ 
\cline{7-10} &&&&&&TC$_{easy}$&TC$_{med}$&TC$_{hard}$&TC$_{all}$&&&\\\hline 

DFSC-all& 68.26& 66.23 & 58.66& 54.60 & 56.96 & 77.38 & 63.65 & 44.14 & 61.12 & 50.51 & 63.99 & 67.94\\
\rowcolor{mygray}DFSC-cls&70.22&\textbf{68.47}&60.78&55.49&60.02&\textbf{78.17}&65.75&45.07&62.60&52.29&\textbf{66.12}&69.53\\
DFSC-reg&\textbf{70.31}&68.26&60.70&55.31&59.05&77.52&65.52&\textbf{46.74}&\textbf{62.82}&51.76&66.06&68.95\\
\rowcolor{mygray}DFSC& 70.16 & 68.07 & \textbf{60.95} & \textbf{55.55} & \textbf{60.13} & 77.96 & \textbf{65.85} & 45.59 & 62.75 & \textbf{53.10} & 66.04 & \textbf{69.84} \\
\Xhline{1.2pt}
\hline\hline
\Xhline{1.2pt}

DFSC-$\alpha$0.25&\textbf{70.63}&\textbf{69.18}&60.80&\textbf{55.74}&59.61&\textbf{78.56}&65.70&\textbf{46.26}&\textbf{63.00}&52.75&\textbf{66.33}&\textbf{70.31}\\
\rowcolor{mygray}DFSC-$\alpha$0.5&70.37& 68.32&60.20&55.01&58.86&77.17&64.49&45.28&61.82&51.52&65.25&69.55\\
DFSC-$\alpha$1& 70.16 & 68.07 & \textbf{60.95} & 55.55 & \textbf{60.13} & 77.96 & \textbf{65.85} & 45.59 & 62.75 & \textbf{53.10} & 66.04 & 69.84 \\

\rowcolor{mygray}DFSC-$\alpha$2&69.76& 68.19&60.46&55.20&59.53&77.57&65.43&45.42&62.41&52.85&65.61&69.98\\
\Xhline{1.2pt}
\hline \end{tabular} \label{tab:ourtest} \end{table*}

\setlength{\tabcolsep}{9pt}
\begin{table}
\begin{center}
\caption{Comparison of different training strategy in terms of precision (\%) and success (\%) on Anti-UAV. }
\label{tab:sp}
\begin{tabular}{c||c|c|c|c} \hline\Xhline{1.2pt}
\multirow{2}{*}{\textbf{Method}} & \multicolumn{2}{c|}{\textbf{Infrared}} & \multicolumn{2}{c}{\textbf{Visible}} \\
\cline{2-5}
&Precision&Success&Precision&Success\\ \hline\Xhline{1.2pt}
\rowcolor{mygray}large-scale & 85.34&61.13&88.53&63.00\\
normal& 87.61&62.88&93.30&65.22\\
\rowcolor{mygray}DFSC& 87.77&63.24&\textbf{93.87}&65.87\\
\hline
\Xhline{1.2pt}
DFSC-all& 86.97&61.59&91.87&64.20\\
\rowcolor{mygray}DFSC-cls& 87.47&63.14&93.72&65.56\\
DFSC-reg& \textbf{87.87}&63.35&93.26&65.24\\
\hline
\Xhline{1.2pt}
\rowcolor{mygray}DFSC-$\alpha$0.25&87.52&\textbf{63.40}&93.65&\textbf{66.24}\\
DFSC-$\alpha$0.5&87.08&62.53&93.84&65.66\\
\rowcolor{mygray}DFSC-$\alpha$1&87.77&63.24&93.83&65.50\\
DFSC-$\alpha$2&87.83&62.81&93.74&65.94\\
\hline\Xhline{1.2pt}
\end{tabular}
\end{center}
\vspace{-7mm}
\end{table}

\begin{figure*}[h]
\begin{center}
\includegraphics[width=1\linewidth]{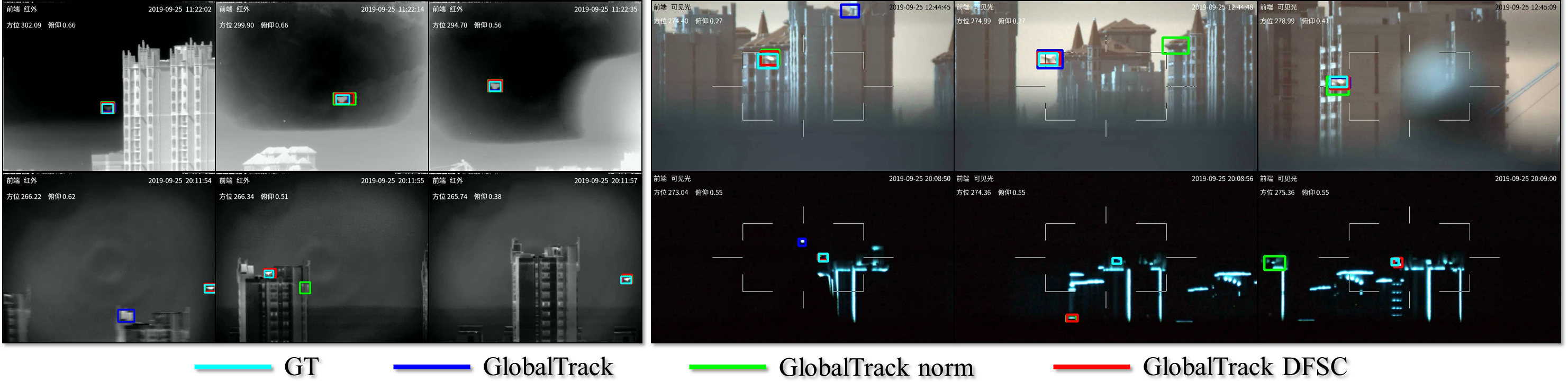}
\end{center}
 \vspace{-4mm}
 \caption{A comparison of proposed DFSC with other method. Through the visualization of successful tracking sequences, it is shown that DFSC successfully handles these challenges and provides accurate UAV state estimation. Best viewed in color. }
\label{fig:final_example}
\vspace{-3mm}
\end{figure*}

\vspace{-3mm}
\subsection{Evaluation under Protocol II}
\label{sec:define}
Here, the finetuning of the GlobalTrack model only with Anti-UAV directly is defined as the normal training strategy. Otherwise, finetuning the GlobalTrack model using the DFSC method only with Anti-UAV is called the DFSC training strategy. The model provided on GitHub is called the large-scale training strategy, which utilizes MS COCO, LaSOT, and GOT-10k as the training set.

\noindent\textbf{Overall Performance.} As shown at the top of Tab. \ref{tab:ourval}, DFSC obtains the best performance on both infrared and visible sequences. Compared with the normal training strategy, DFSC gets a gains a 0.49 $mSA$ gain on the validation set and 0.68 $mSA$ gain on the test set. For visible tracking sequence, DFSC improves $mSA$ by 0.48 on the validation set and 0.57 on the test set, respectively.
And compared with large-scale training strategy, the normal and DFSC methods gain obviously on validation set. This is because, in Anti-UAV, the sequences in training set and validation set may come from the same video. Through the learning of similar data, the tracker can make a more accurate judgment. However, since test set is independent of the other two sets, the performance gains of normal and DFSC method decrease on test set, showing that the tracker is overfitting in the training set to some extent.

Similar observations can be obtained in the metric comparison among different methods on precision and success as shown in Tab. \ref{tab:sp}, which demonstrates that the proposed DFSC gets consistent performance improvement. 

\noindent\textbf{Attribute-based Performance.} 
Different training strategies are evaluated on the defined attribute annotations to analyze the advantages and shortcomings of the proposed method on specific attributes. Tab. \ref{tab:ourtest} shows that the proposed method improves the performance of most attributes on the validation set, especially OC. Compared with the normal training strategy, DFSC gains the $mSA$ by 1.28\%, 0.97\% and 0.78\% on OC, SV and LR, respectively. In other attributes, performance improvement is limited. 

Moreover, for the test set, DFSC gets the best $mSA$ score (62.75\%) on $\rm TC_{all}$, which is higher than the normal training strategy by 1.07\%. Compared with normal training strategy, DFSC gains the $mSA$ by 0.89\%, 1.21\% and 0.98\% and on $\rm TC_{easy}$, $\rm TC_{med}$ and $\rm TC_{hard}$, respectively.

On the test set, the large-scale training strategy achieves the highest performance in most attributes except TC and OV. However, the whole performance is not as good as DFSC on $mSA$. On the one hand, the large-scale training strategy uses more data than the normal and DFSC method, which is conducive to learning more general features beneficial to tracking UAVs. On the other hand, after training on the Anti-UAV training set, the tracker has a more discrimination ability on TC scenarios and is more stable in common scenarios.

To validate the proposed DFSC method, the successful tracking sequences from test set are visualized, as shown in Fig. \ref{fig:final_example}. As a consequence, a potential training strategy is demonstrated to handle such great challenges.

\vspace{-4mm}
\subsection{Ablation Study}
This subsection provides the analysis about the behaviors of the supervised task and $\alpha$ in Eq. \ref{CSM} for the dual-flow semantic modulation (DFSC) training strategy.

\noindent\textbf{Supervised Task.} As shown at the middle of Tab. \ref{tab:ourtest}, compared with other methods, DFSC-cls achieves the best performance on $mSA$, but has the lowest performance on success and precision on the infrared dataset. Nevertheless, DFSC-cls only gets a small margin over DFSC. Instead, DFSC-reg obtains a lower $mSA$ but the highest precision and success scores. DFSC-cls makes the tracker own a more accurate UAV state discrimination ability, but its localization ability is not improved relatively. It validates the classification supervised task in the DFSC method to force the tracker to focus on the powerful capability of UAV perception. On the contrary, more attention to the regression task makes the tracker perform better on UAV localization.

Compared with the DFSC method, DFSC-all both adopt cross sequence semantic information to modulate features in the CSM and ISM stage. In both evaluations, DFSC performs favorably compared with DFSC-all. Class-level semantic modulation in the later stage will make the tracker confused about the information from different UAVs. As a result, instance-level semantic modulation is necessary for the tracker to enhance discrimination power.

\noindent\textbf{Influence of Ratio.} The $\alpha$ means the ratio between $L_{cross}$ and $L_{same}$ in Eq. \ref{CSM}. As illustrated in the bottom of Tab. \ref{tab:ourtest}, $mSA$ sharply fluctuates if the ratio $\alpha$ changes from large to small, in which case DFSC becomes similar to normal method. 

With the increase of the ratio $\alpha$, the performance first increases and then decreases. When the performance reaches the maximum, $\alpha$ is about 0.25. And if $\alpha$ becomes smaller or larger than this value, the performance will decline. This phenomenon is because when $\alpha$ becomes too large, the tracker will unnecessarily focus too much on the cross sequence feature information. It hinders the tracker from learning class-level semantic feature, which reduces the robustness of the tracker. When $\alpha$ becomes smaller, DFSC will degenerate into 
the normal training strategy. A moderate ratio ($\alpha$=0.25) can achieve a trade-off between intra-sequence semantics and cross-sequence semantics.

\vspace{-3mm}
\section{Conclusion}
In the community, there is no high-quality Anti-UAV benchmark for capturing real dynamic scenes. In this paper, the first UAV tracking dataset is constructed, named Anti-UAV, which collects over 300 video pairs and annotates more than 580k bounding boxes manually. Along with the dataset, evaluation protocols, metrics, and baseline trackers are introduced for the task of UAV tracking. Furthermore, a novel approach named dual-flow semantic consistency (DFSC) is proposed for UAVs tracking. DFSC enables the tracker to fully leverage the semantic information across different video sequences, such that tracker's robustness and discrimination ability can be further improved. Notably, the proposed DFSC does not introduce any additional inference time. In the future, multi-modal with unaligned data for tracking will be further investigated, which has the potential to boost the accurateness of trackers. 

\ifCLASSOPTIONcaptionsoff
\newpage
\fi

\footnotesize
\bibliographystyle{IEEEtran}
\bibliography{IEEEabrv,egbib}

\vspace{-13mm}
\begin{IEEEbiography}[{\includegraphics[width=1in,height=1.25in,clip,keepaspectratio]{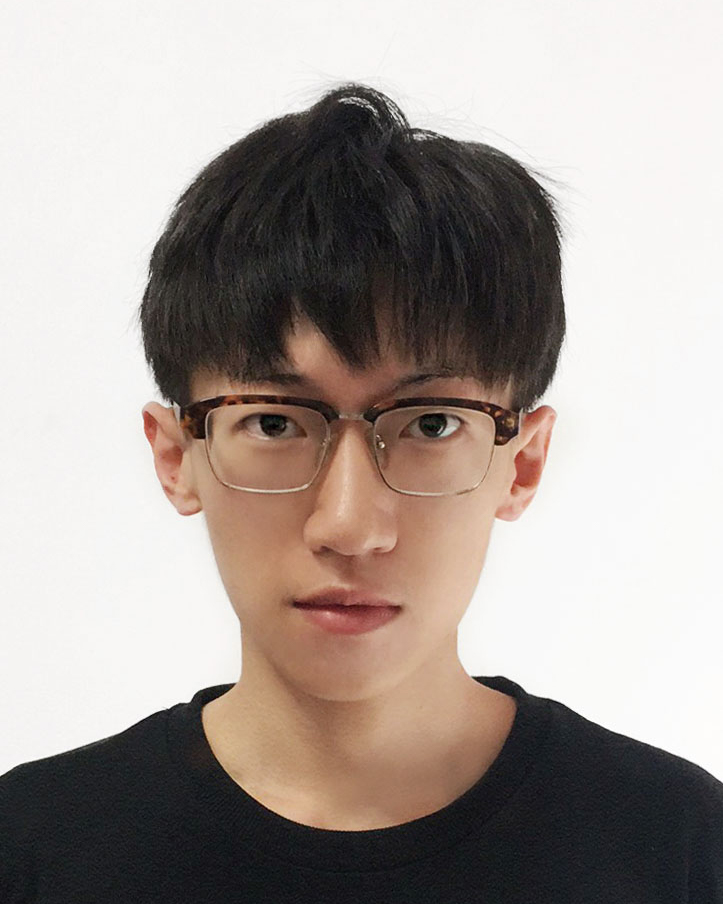}}]{Nan Jiang}
received the B.E. degree in communication engineering from Xi'an Jiaotong University, China, in 2018. He is currently pursuing the M.S. degree in computer science with University of Chinese Academy of Sciences. His research interests include machine learning and computer vision.
\end{IEEEbiography}

\vspace{-13mm}

\begin{IEEEbiography}[{\includegraphics[width=1in,height=1.25in,clip,keepaspectratio]{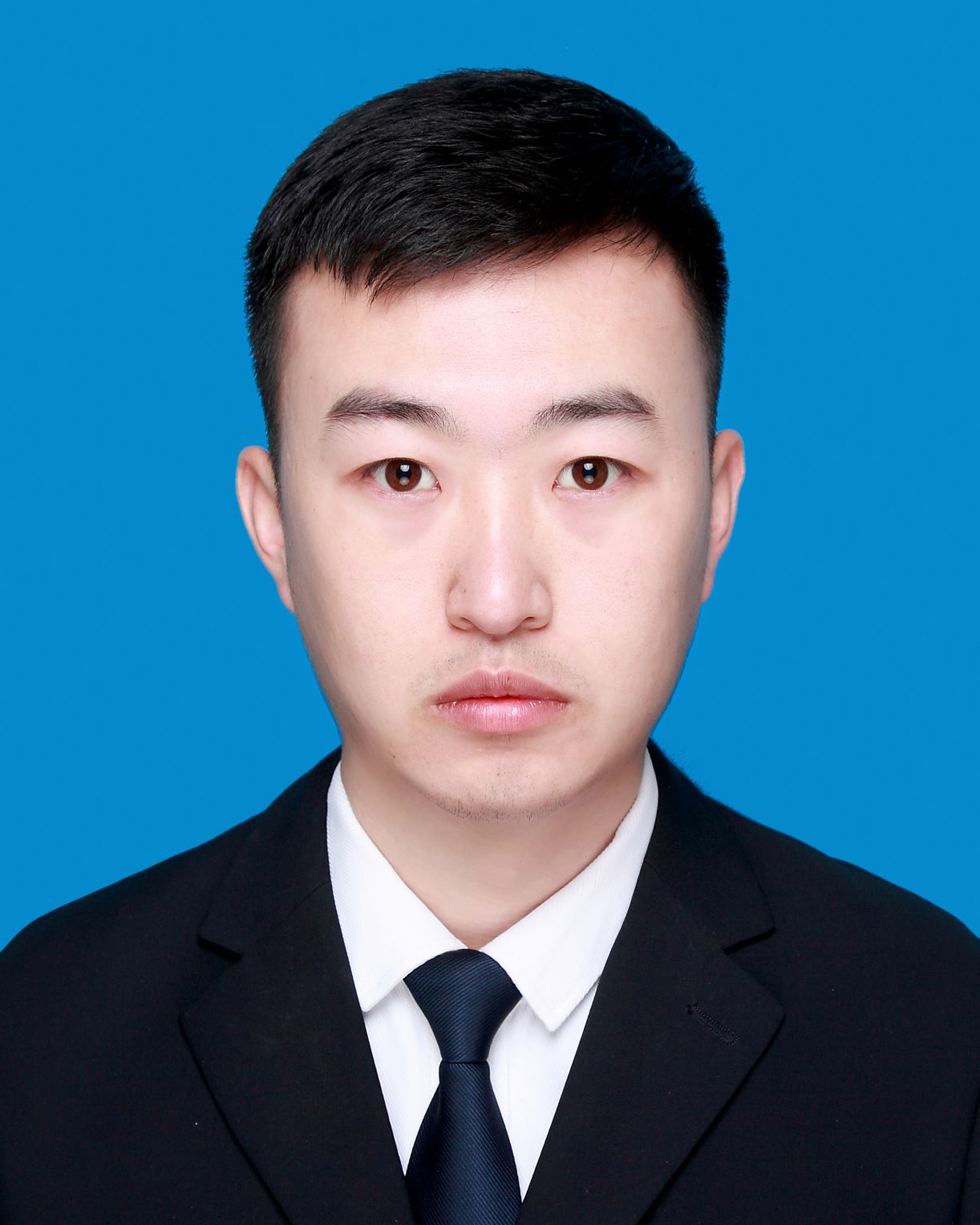}}]{Kuiran Wang}
 received the B.E. degree in computer science and technology from Central South University, China, in 2018. He is currently pursuing the M.S. degree in electronic and communication engineering with University of Chinese Academy of Sciences. His research interests include machine learning and computer vision.
\end{IEEEbiography}

\vspace{-13mm}

\begin{IEEEbiography}[{\includegraphics[width=1in,height=1.25in,clip,keepaspectratio]{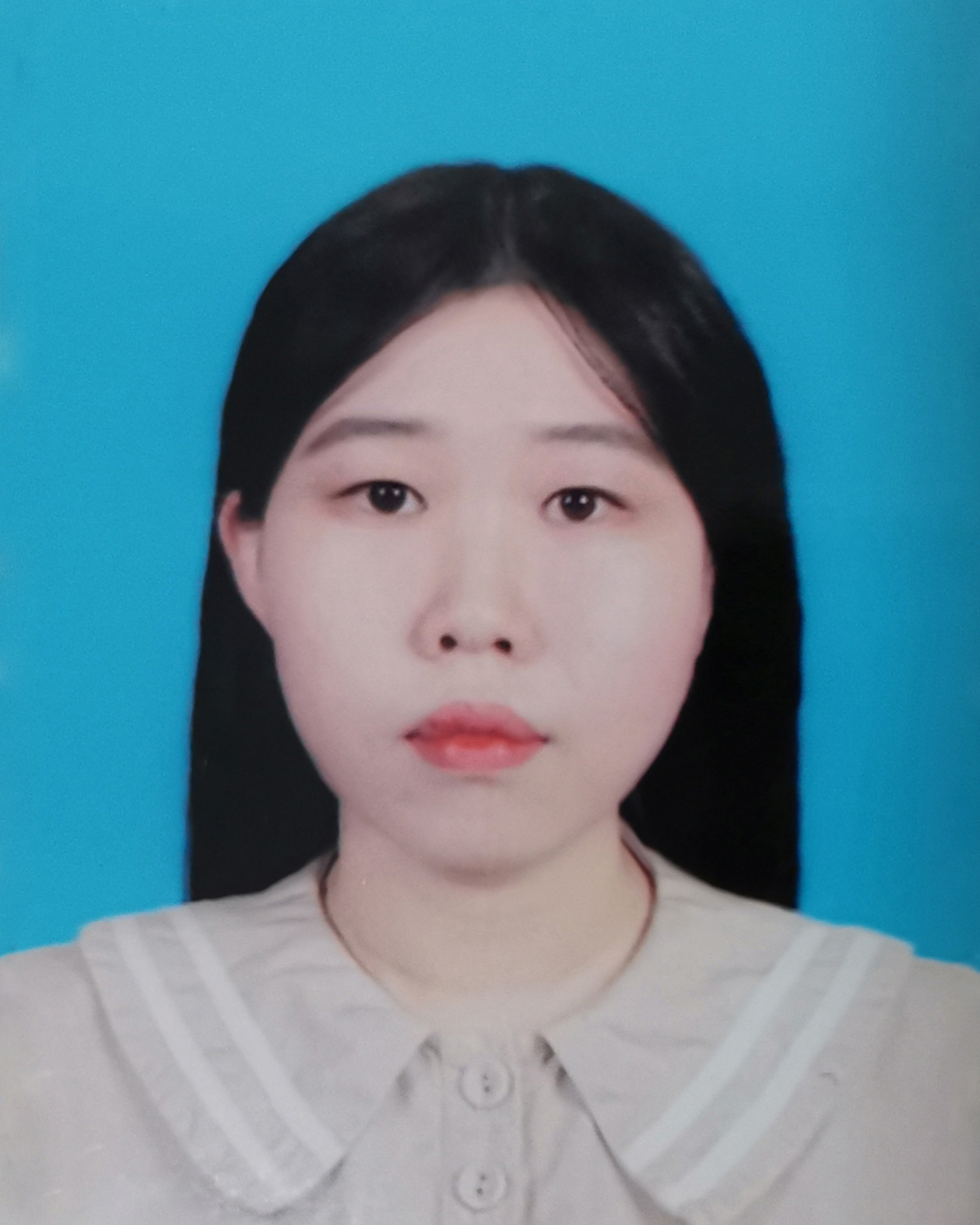}}]{Xiaoke Peng}
received the B.E. degree in communication engineering from Sun Yat-sen University, China, in 2019. She is currently pursuing the M.S. degree in electronic and communication engineering with University of Chinese Academy of Sciences. Her research interests include machine learning and computer vision.
\end{IEEEbiography}

\vspace{-13mm}

\begin{IEEEbiography}[{\includegraphics[width=1in,height=1.25in,clip,keepaspectratio]{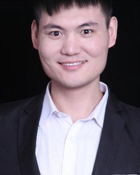}}]{Xuehui Yu}
received the B.E. degree in software engineering from Tianjin University, China, in 2017. He is currently pursuing the Ph.D. degree in signal and information processing with University of Chinese Academy of Sciences. His research interests include machine learning and computer vision.
\end{IEEEbiography}

\vspace{-13mm}

\begin{IEEEbiography}[{\includegraphics[width=1in,height=1.25in,clip,keepaspectratio]{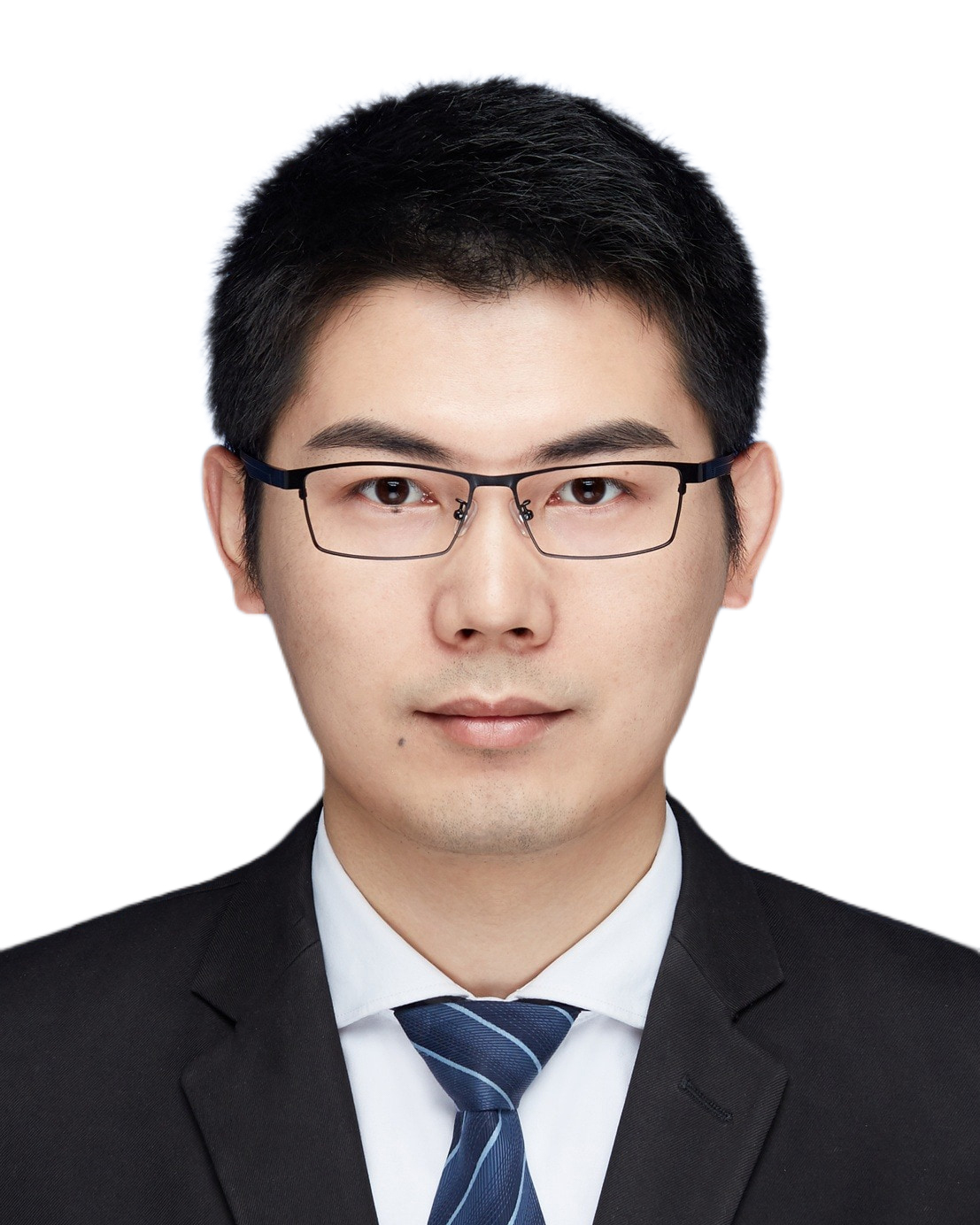}}]{Qiang Wang}
received the B.S. degree from the University of Science and Technology Beijing, Beijing, China, in 2015, and the Ph.D. degree from the University of Chinese Academy of Sciences (UCAS) in 2020. His research interests include visual tracking, autonomous vehicles, and service robots.
\end{IEEEbiography}

\vspace{-13mm}

\begin{IEEEbiography}[{\includegraphics[width=1in,height=1.25in,clip,keepaspectratio]{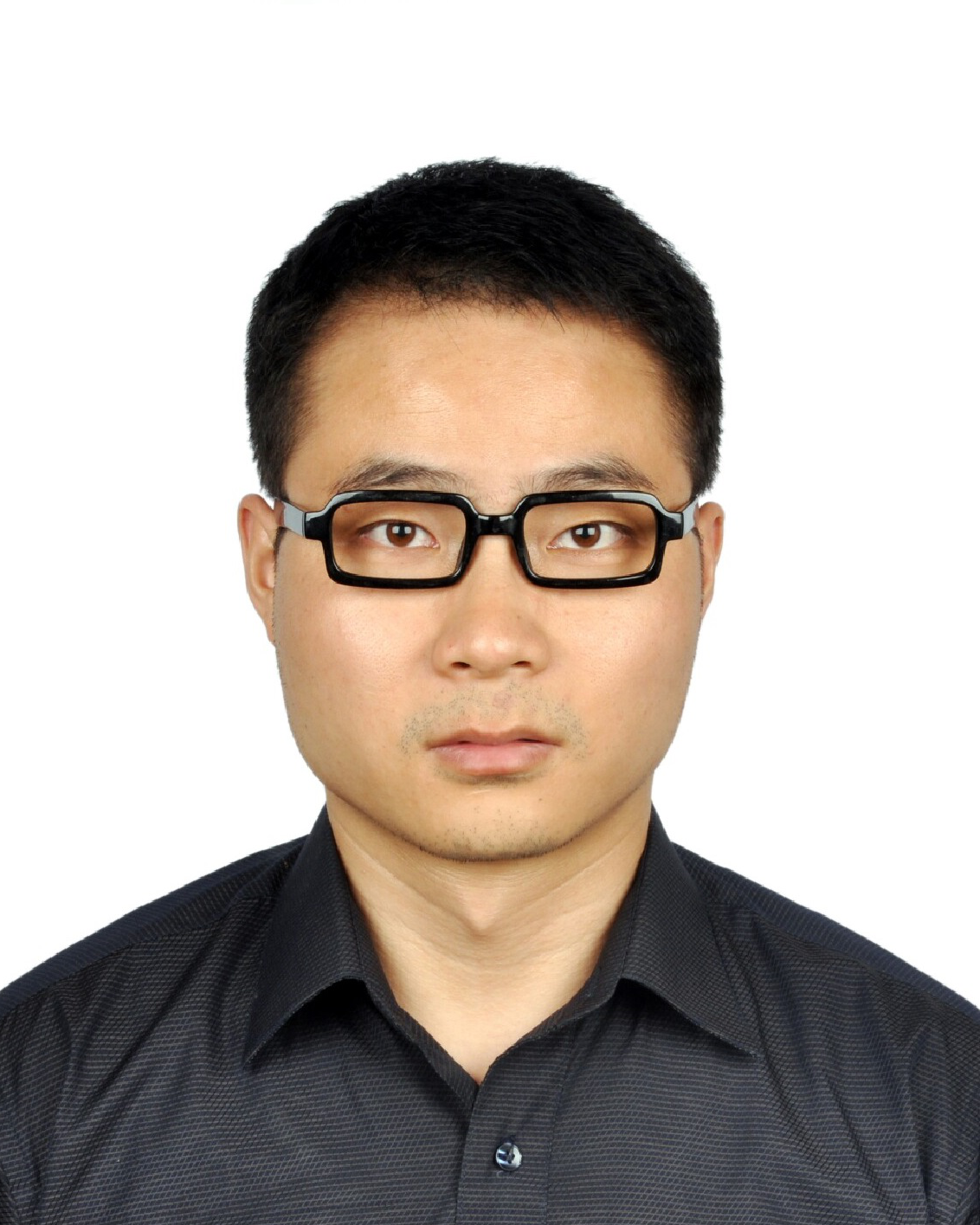}}]{Junliang Xing}
received his dual B.S. degrees in computer science and mathematics from Xi'an Jiaotong University, Shanxi, China, in 2007, and the Ph.D. degree in computer science from Tsinghua University, Beijing, China, in 2012. He is currently a Professor with the National Laboratory of Pattern Recognition, Institute of Automation, Chinese Academy of Sciences, Beijing, China. Dr. Xing was the recipient of Google Ph.D. Fellowship 2011, the Excellent Student Scholarships at Xi'an Jiaotong University from 2004 to 2007 and at Tsinghua University from 2009 to 2011. He has published more than 100 papers on international journals and conferences. His current research interests mainly focus on computer vision problems related to faces and humans.
\end{IEEEbiography}

\vspace{-5mm}

\begin{IEEEbiography}[{\includegraphics[width=1in,height=1.25in,clip,keepaspectratio]{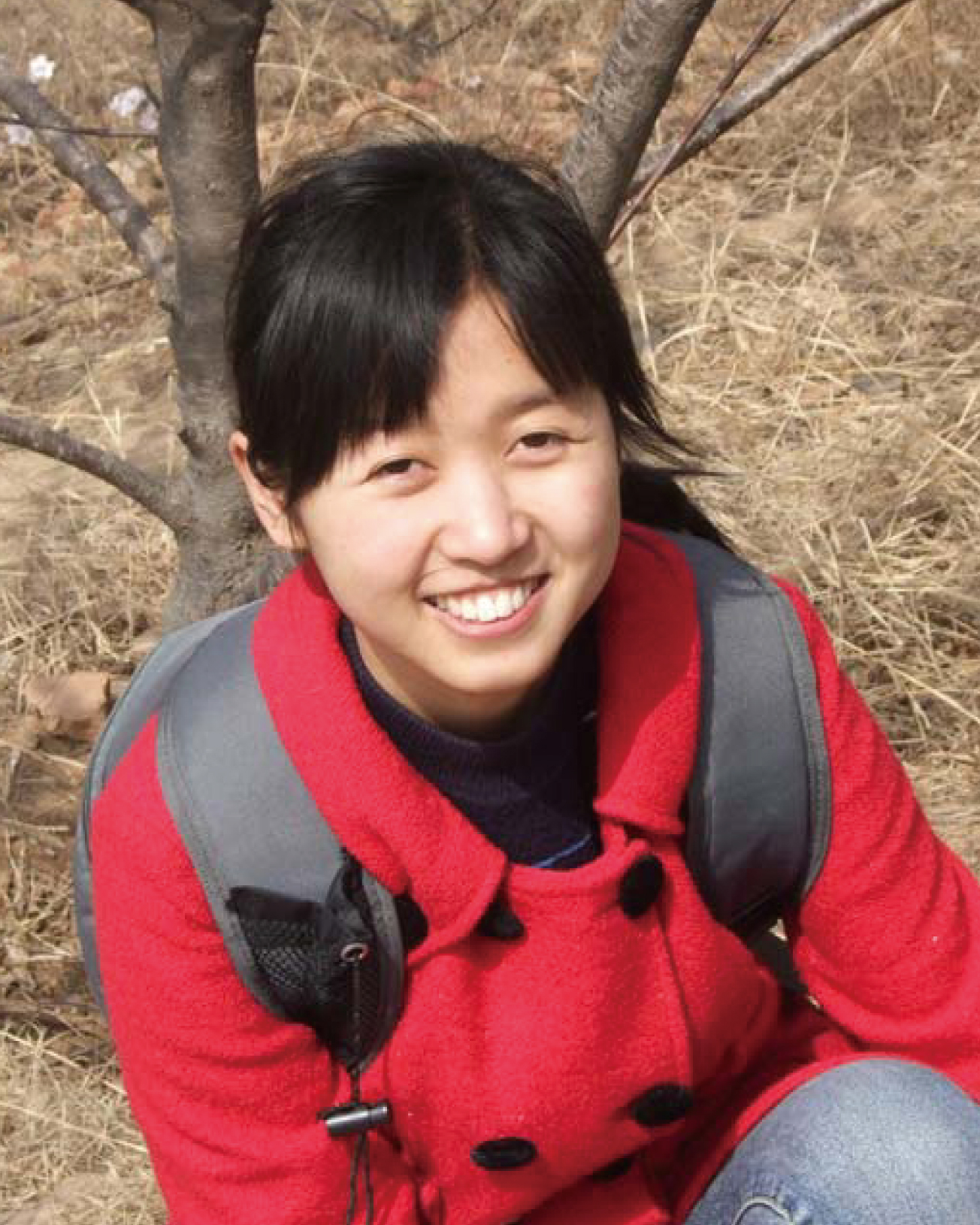}}]{Guorong Li}
received her B.S. degree in technology of computer application from Renmin University of China, in 2006 and Ph.D. degree in technology of computer application from the Graduate University of the Chinese Academy of Sciences in 2012. 
\par Now, she is an associate professor at the University of Chinese Academy of Sciences. Her research interests include object tracking, video analysis, pattern recognition, and cross-media analysis.
\end{IEEEbiography}

\vspace{-10mm}

\begin{IEEEbiography}[{\includegraphics[width=1in,height=1.25in,clip,keepaspectratio]{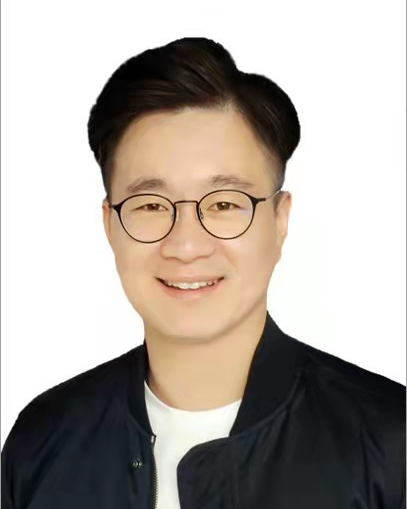}}]{Jian Zhao}
received the Bachelors degree from Beihang University in 2012, the Masters degree from National University of Defense Technology in 2014, and the Ph.D. degree from National University of Singapore in 2019. He is currently an Assistant Professor with Institute of North Electronic Equipment, Beijing, China. His main research interests include deep learning, pattern recognition, computer vision and multimedia analysis. He has published over 40 cutting-edge papers. He has received ``2021-2023 Beijing Young Talent Support Project" from Beijing Association for Science and Technology in 2020. He has won the Lee Hwee Kuan Award (Gold Award) on PREMIA 2019, the “Best Student Paper Award on ACM MM 2018, and the top-3 awards several times on world-wide competitions. He is the EAC of VALSE, and the committee member of CSIGBVD. He has served as the invited reviewer of NSFC, T-PAMI, IJCV, T-MM, T-IFS, T-CSVT, Neurocomputing, T-CDS, CSSP, JVCI, NeurIPS (one of the top 30\% highest-scoring reviewers of NeurIPS 2018), CVPR, ICCV, ACM MM, AAAI, ICLR, ICML, ACCV, UAI.
\end{IEEEbiography}

\vspace{-10mm}

\begin{IEEEbiography}[{\includegraphics[width=1in,height=1.25in,clip,keepaspectratio]{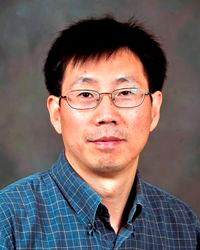}}]{Guodong Guo}
received the B.E. degree in Automation from Tsinghua University, Beijing, China,
the Ph.D. degree in Computer Science from University
of Wisconsin, Madison, WI, USA. He is
the Head of the Institute of Deep
Learning, Baidu Research, and also affiliated with  
the Dept. of Computer Science
and Electrical Engineering, West Virginia University
(WVU), USA. In the past, he visited and worked in
several places, including INRIA, Sophia Antipolis,
France; Ritsumeikan University, Kyoto, Japan; and
Microsoft Research, Beijing, China; He authored a book, “Face, Expression,
and Iris Recognition Using Learning-based Approaches” (2008), co-edited
two books, “Support Vector Machines Applications” (2014) and “Mobile
Biometrics” (2017), and co-authored a book, “Multi-Modal Face Presentation
Attack Detection” (2020). He published over 150 technical papers. His
research interests include computer vision, biometrics, machine learning, and
multimedia. He is an AE of several journals, including IEEE Trans. on Affective Computing.He received the North Carolina State Award for Excellence
in Innovation in 2008, Outstanding Researcher (2017-2018, 2013-2014) at
CEMR, WVU, and New Researcher of the Year (2010-2011) at CEMR, WVU.
He was selected the “People’s Hero of the Week” by BSJB under Minority
Media and Telecommunications Council (MMTC) in 2013. Two of his papers
were selected as “The Best of FG’13” and “The Best of FG’15”, respectively.

\end{IEEEbiography}

\vspace{-10mm}

\begin{IEEEbiography}[{\includegraphics[width=1in,height=1.25in,clip,keepaspectratio]{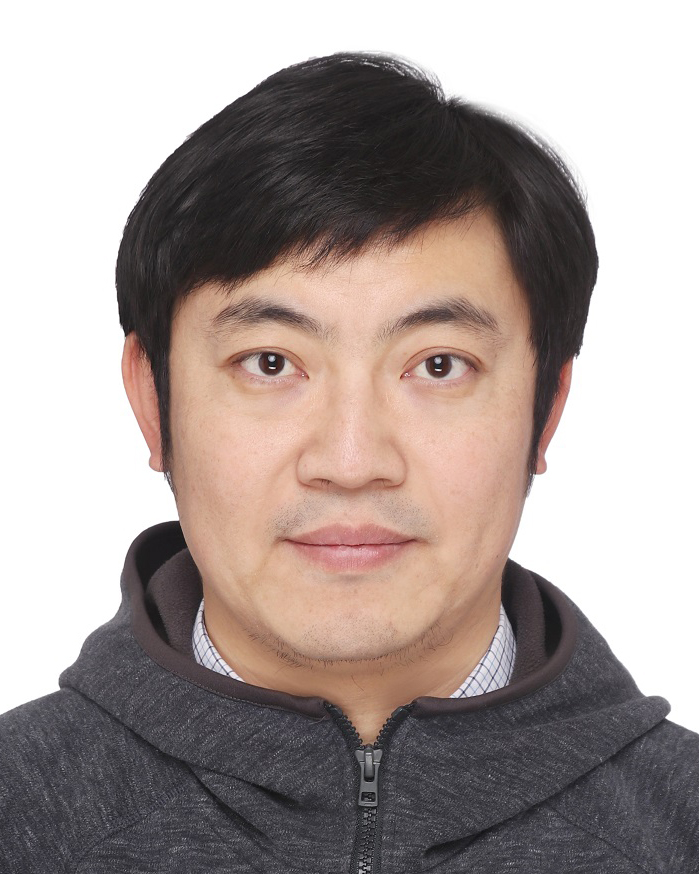}}]{Zhenjun Han}
received the B.S. degree in software engineering from Tianjin University, Tianjin, China, in 2006 and the M.S. and Ph.D. degrees from University of Chinese Academy of Sciences, Beijing, China, in 2009 and 2012, respectively. Since 2013, he has been an Associate Professor with the School of Electronic, Electrical, and Communication Engineering, University of Chinese Academy of Sciences. His research interests include object tracking and detection.
\end{IEEEbiography}

\end{document}